\documentclass[10pt,journal,compsoc]{IEEEtran}
\usepackage{xcolor}
\usepackage{booktabs}   
\usepackage{multirow}  
\usepackage{colortbl}  
\usepackage{arydshln}  
\usepackage{graphicx}  
\usepackage{tabularx}
\usepackage{cite}
\usepackage{url}
\usepackage{amsmath}
\usepackage{amssymb}
\usepackage[caption = false, font=footnotesize]{subfig}
\usepackage{float}
\usepackage{booktabs} 
\usepackage{multirow}
\usepackage{array}
\usepackage{color}
\usepackage{longtable}
\usepackage{tabu}
\usepackage{bbding}
\usepackage{pifont}
\usepackage{hyperref} 
\usepackage{enumitem}

\newcolumntype{L}[1]{>{\raggedright\let\newline\\\arraybackslash\hspace{0pt}}m{#1}}
\newcolumntype{C}[1]{>{\centering\let\newline\\\arraybackslash\hspace{0pt}}m{#1}}
\newcolumntype{R}[1]{>{\raggedleft\let\newline\\\arraybackslash\hspace{0pt}}m{#1}}
\usepackage{algorithm, algorithmic}
\newcommand{\etal}{\textit{et al}.~}
\newcommand{\ie}{\textit{i}.\textit{e}.}
\newcommand{\eg}{\textit{e}.\textit{g}.}

\newcommand{\br}[1]{{\color{red}#1}}
\newcommand{\gr}[1]{{\color{blue}#1}}

\definecolor{lightcyan}{RGB}{180,240,240}
\newcommand{\rc}[1]{{\color{black}#1}}
\hypersetup{
    colorlinks,
    linkcolor={blue},
    urlcolor={magenta}
}

\begin{document}

\title{From Global to Granular: Revealing IQA Model Performance via Correlation Surface}

\author{Baoliang Chen,
        Danni Huang,
        Hanwei Zhu,
        Lingyu Zhu,
        Wei Zhou,~\IEEEmembership{Senior Member,~IEEE}
        Shiqi Wang,~\IEEEmembership{Senior Member,~IEEE}
        Yuming Fang,~\IEEEmembership{Fellow,~IEEE}
        and~Weisi Lin,~\IEEEmembership{Fellow,~IEEE}% <-this % stops a space
\IEEEcompsocitemizethanks{
\IEEEcompsocthanksitem Baoliang Chen is with the School of Computer Science, South China Normal University, and the College of Computing and Data Science, Nanyang Technological University. E-mail: blchen6-c@my.cityu.edu.hk.
\IEEEcompsocthanksitem Danni Huang is with the School of Computer Science, South China Normal University. E-mail: dannyhuang@m.scnu.edu.cn.

\IEEEcompsocthanksitem  Lingyu Zhu and Shiqi Wang are with the School of Computer Science, City University of Hong Kong. (E-mails: lingyzhu-c@my.cityu.edu.hk, shiqwang@cityu.edu.hk.)
\IEEEcompsocthanksitem Wei Zhou is with the School of Computer Science and Informatics, Cardiff University, Cardiff, CF24 4AG, United Kingdom. (E-mail:zhouw26@cardiff.ac.uk);
\IEEEcompsocthanksitem Yuming Fang is
with the School of Computing and Artificial Intelligence, Jiangxi University
of Finance and Economics, Nanchang, China, and also with Jiangxi Provincial Key Laboratory of Multimedia Intelligent Processing. (E-mail: fa0001ng@e.ntu.edu.sg).
\IEEEcompsocthanksitem Hanwei Zhu, and W. Lin are with the College of Computing and Data Science, Nanyang Technological University. (E-mails: hanwei.zhu@ntu.edu.sg, wslin@ntu.edu.sg.)
\IEEEcompsocthanksitem This work was supported in part by the National Natural Science Foundation of China under Grants 62401214, U24A20220, and 62132006, and in part by the Ministry of Education, Singapore, under Tier 1 Grant RG103/24. Corresponding author: Hanwei Zhu.
}}

\markboth{Submitted to IEEE Transactions on Pattern Analysis and Machine Intelligence}{}

\IEEEtitleabstractindextext{%
\begin{abstract}
Evaluation of Image Quality Assessment (IQA) models has long been dominated by global correlation metrics, such as Pearson Linear Correlation Coefficient (PLCC) and Spearman Rank-Order Correlation Coefficient (SRCC). While widely adopted, these metrics reduce performance to a single scalar, failing to capture how ranking consistency varies across the local quality spectrum. For example, two IQA models may achieve identical SRCC values, yet one ranks high-quality images (related to high Mean Opinion Score, MOS) more reliably, while the other better discriminates image pairs with small quality/MOS differences (related to $|\Delta$MOS$|$). Such complementary behaviors are invisible under global metrics. Moreover, SRCC and PLCC are sensitive to test-sample quality distributions,  yielding unstable comparisons across test sets.
To address these limitations, we propose \textbf{Granularity-Modulated Correlation (GMC)}, which provides a structured, fine-grained analysis of IQA performance. GMC includes: (1) a \textbf{Granularity Modulator} that applies Gaussian-weighted correlations conditioned on absolute MOS values and pairwise MOS differences ($|\Delta$MOS$|$) to examine local performance variations, and (2) a \textbf{Distribution Regulator} that regularizes correlations to mitigate biases from non-uniform quality distributions. The resulting \textbf{correlation surface} maps correlation values as a joint function of MOS and $|\Delta$MOS$|$, providing a 3D representation of IQA performance. 
Experiments on standard benchmarks show that GMC reveals performance characteristics invisible to scalar metrics, offering a more informative and reliable paradigm for analyzing, comparing, and deploying IQA models. Codes are available at \url{https://github.com/Dniaaa/GMC}.
\end{abstract}

}

% make the title area
\maketitle
\IEEEdisplaynontitleabstractindextext
\IEEEpeerreviewmaketitle

\section{Introduction}
\label{sec:introduction}

\IEEEPARstart{D}{riven} by the proliferation of visual media and advances in artificial intelligence, image data has witnessed substantial growth in both volume and complexity. This evolution, in turn, has led to a growing demand for accurate image quality assessment (IQA), which plays a fundamental role in image compression, enhancement, restoration, and generation\cite{wang2006modern,zhai2020perceptual,ding2021comparison}. Over the past decades, a variety of objective Full-Reference (FR) and No-Reference (NR) IQA models have been proposed, ranging from perceptual fidelity\cite{mannos1974effects, ding2020image, wang2009mean, girod1993s} to data-driven paradigms\cite{zhang2018blind, kim2018deep, bosse2017deep, chen2024tip, you2021transformer, cheon2021perceptual, ke2021musiq, zhang2023blind, xu2024boosting, wu2023qbench, wu2025visualquality}. However, as IQA models become increasingly sophisticated, a critical question arises:\\

\noindent \textit{How should we comprehensively and fairly measure the performance of these IQA models?}\\

Historically, the IQA community has relied on global correlation metrics, such as the Pearson Linear Correlation Coefficient (PLCC), Spearman Rank-Order Correlation Coefficient (SRCC), and Kendall Rank-Order Correlation Coefficient (KRCC), to quantify the alignment between model predictions and Mean Opinion Scores (MOS). However, with the increasing diversification of IQA applications, these global indicators are becoming inadequate for capturing the nuanced demands of real-world scenarios. For instance, in \textbf{generative AI}, where the majority of outputs possess high perceptual quality, there is a specific need for models that are highly accurate in the high-MOS regime. Conversely, in \textbf{extreme image compression} settings, models are expected to perform fine-grained discrimination among images with marginal differences in perceived quality (small $|\Delta \mathrm{MOS}|$), which is crucial for effective rate–distortion optimization and requires strong fine-grained differentiation capability \cite{zhang2021fine}. In practice, two IQA models might exhibit nearly identical global SRCC scores on a standard benchmark, yet their performance could diverge significantly in these specialized application contexts.

\begin{figure*}[htbp] 
\centering 
\includegraphics[width=\textwidth]{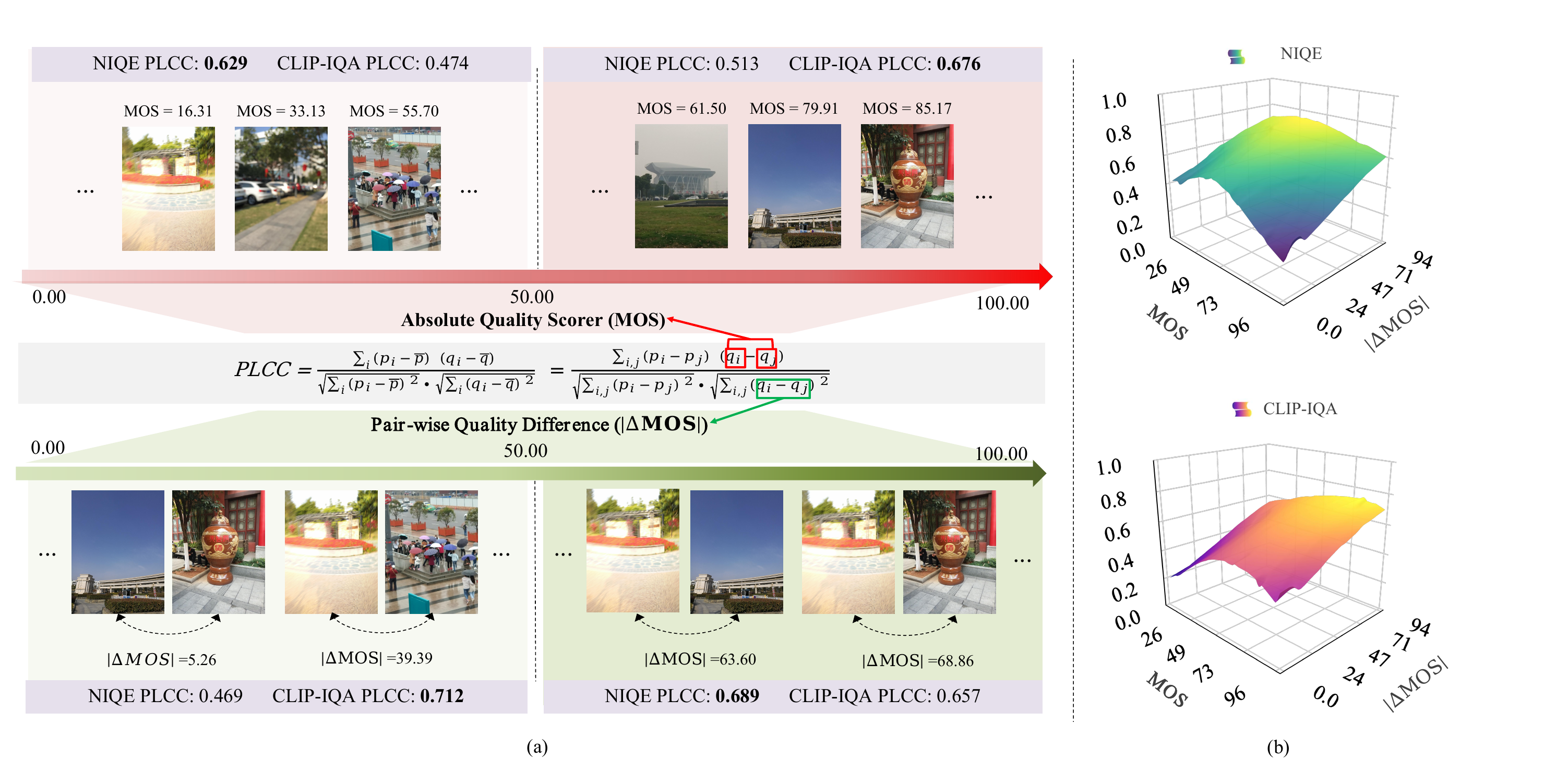} 
\caption{\textbf{Complementary IQA behaviors along two coupled assessment dimensions on the SPAQ dataset~\cite{fang2020perceptual}.} 
The overall performance of an IQA model is governed by its \textit{prediction accuracy} relative to absolute quality (MOS) and its \textit{discrimination capability} regarding pairwise differences ($|\Delta\text{MOS}|$). 
(a) Performance snapshots: Images from the SPAQ dataset are partitioned into four subsets based on the medians of MOS and $|\Delta\text{MOS}|$. PLCC scores on these quadrants reveal the distinct regimes where models excel. For instance, CLIP-IQA~\cite{wang2023exploring} is proficient in high-quality and fine-grained discrimination, while NIQE~\cite{mittal2012making} is more robust for severe degradations. 
(b) Correlation surfaces estimated by the proposed GMC provide a unified and comprehensive view, capturing how correlation performance (\eg, PLCC) evolves seamlessly across the joint MOS and $|\Delta\text{MOS}|$ space.}

\label{fig:abs} 
\end{figure*}

\begin{figure}[!t] 
\centering 
\includegraphics[width=0.5\textwidth]{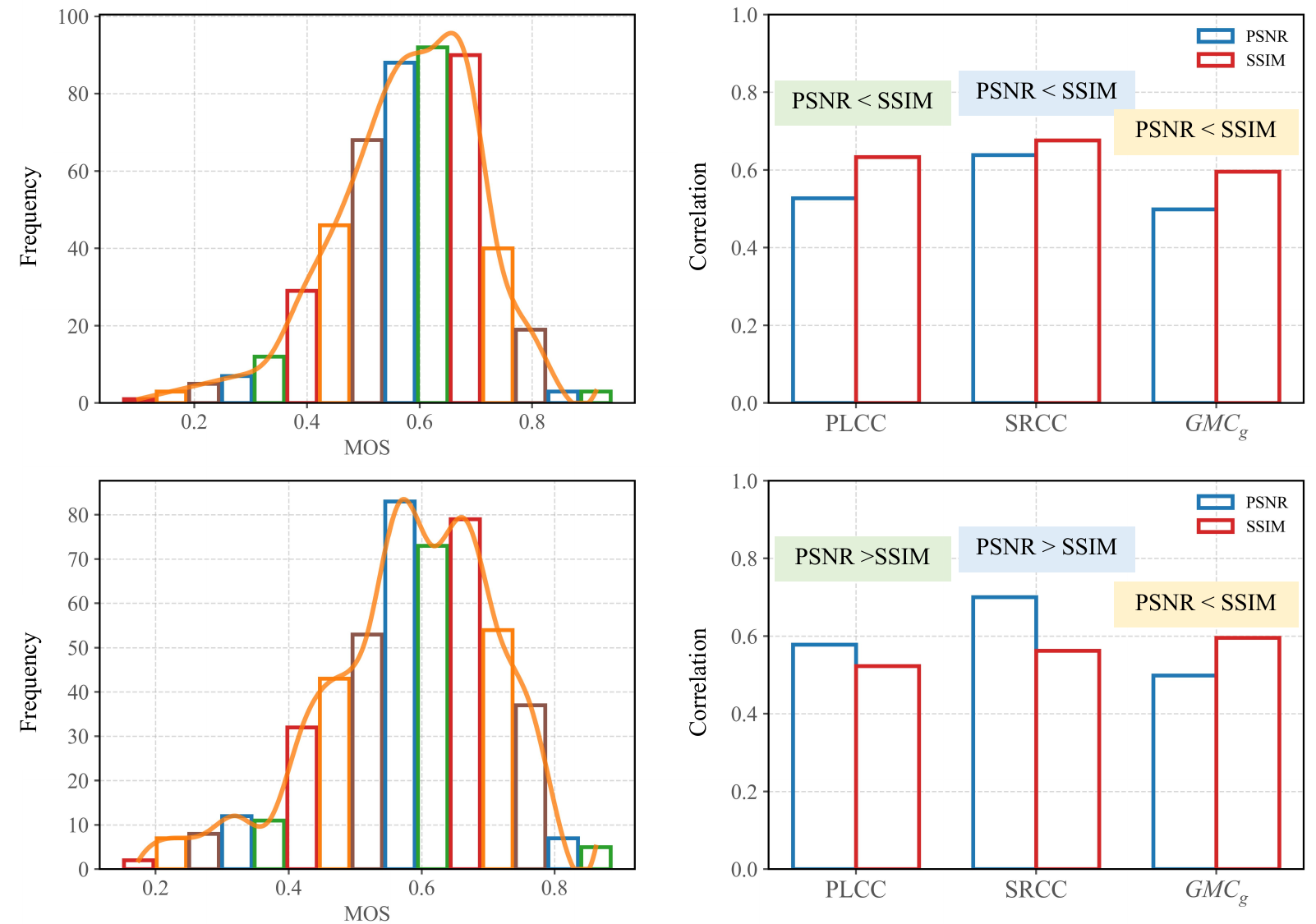} 
\caption{Sensitivity of global correlation metrics (PLCC and SRCC) to shifts in quality score distributions. 
Two subsets are sampled from the PIPAL dataset with identical sample sizes but different MOS distributions. 
The first column shows the MOS distributions of the sampled subsets, while the second column reports the PLCC, SRCC, and the proposed global GMC score ($\mathrm{GMC}_g$) for PSNR and SSIM. 
Although PSNR and SSIM exhibit reversed performance rankings under PLCC and SRCC under distributional shifts, $\mathrm{GMC}_g$ remains stable and preserves consistent model ordering.}

\label{fig:dist} 
\end{figure}
As illustrated in Fig.~\ref{fig:abs}(a), we partition the SPAQ dataset~\cite{fang2020perceptual} into subsets based on absolute quality score (\ie, MOS) and pair-wise quality difference (\ie, $|\Delta\text{MOS}|$), respectively. Our analysis reveals that CLIP-IQA~\cite{wang2023exploring} exhibits superior PLCC performance in high-MOS regions and for pairs with small $|\Delta\text{MOS}|$, suggesting its proficiency in the fine-grained discrimination of high-quality content. In contrast, NIQE~\cite{mittal2012making} demonstrates higher correlation in low-MOS regimes and for pairs with large $|\Delta\text{MOS}|$, indicating its strength in identifying severe degradations. These findings underscore that IQA model performance is jointly governed by absolute quality levels and the granularity of quality differences. \rc{Such complementary behaviors are obscured by traditional global metrics, which aggregate localized performance into a single scalar and thus provide an incomplete characterization of model performance.}
Through this empirical analysis, we identify two fundamental limitations of current benchmarking protocols:
\begin{enumerate}
    \item \textbf{Dimension Confounding:} Global metrics fail to decouple a model's \textit{Prediction Accuracy} (consistency with absolute MOS) from its \textit{Discrimination Capability} (sensitivity to relative $|\Delta\text{MOS}|$). \rc{These two dimensions capture perceptually distinct aspects of human visual assessment, while being mathematically coupled in traditional global metrics.}\\
    \item \textbf{Distributional Bias:} As shown in Fig.~\ref{fig:dist}, global metrics are highly sensitive to the MOS distribution of the test set. Changes in quality span or sampling density can reverse the relative ranking of models, indicating that these metrics confound model capability with dataset-specific distributional bias.
\end{enumerate}
Inspired by these observations, we propose \textbf{Granularity-Modulated Correlation (GMC)}, a principled framework for IQA model evaluation. GMC explicitly modulates correlation computation along both MOS and $|\Delta\text{MOS}|$ dimensions, enabling localized behavioral analysis while reducing bias induced by non-uniform quality distributions. Specifically, GMC extends the classical Generalized Correlation Coefficient (GCC)~\cite{kendall1990j} via two complementary mechanisms:

\begin{itemize}
    \item \textbf{Granularity Modulator:} \rc{Unlike traditional metrics that assign a uniform weight to every image 
pair regardless of their absolute quality level (MOS) or pairwise quality difference 
($|\Delta\text{MOS}|$), our Granularity Modulator employs explicit Gaussian-based 
weighting $w_k^{ij}$ conditioned on target query coordinates $(Q_k^s, Q_k^d)$, 
enabling localized and controllable assessment of model behavior across specific 
perceptual regimes.}\\
    
    \item \textbf{Distribution Regulator:} To address the instability shown in Fig.~\ref{fig:dist}, this regulator employs a kernel-smoothed density estimator to suppress the dominance of overrepresented quality regions while compensating for sparsely sampled intervals. This ensures that the evaluation remains fair and stable, effectively decoupling model assessment from specific quality distributions.
\end{itemize}

\noindent Beyond providing localized snapshots, our framework synthesizes these measurements into a \textbf{unified 3D correlation surface} (see Fig.~\ref{fig:abs}(b)). The surface fitting offers several distinct advantages. 
First, it provides \textbf{enhanced interpretability} by pinpointing exactly where a model fails or excels, allowing researchers to diagnose whether a low global score stems from poor absolute accuracy or a lack of discriminative sensitivity. 
Second, the localized view offers \textbf{task-specific insights}. For example, a model’s high performance in the ``high-quality'' and ``small difference'' regime (as seen in CLIP-IQA) directly validates its suitability for generative AI evaluation, whereas global metrics would obscure this specialized capability. 
\rc{Finally, by mapping performance onto a continuous landscape, we estimate the evolution of model behavior 
while alleviating the severe boundary effects caused by discrete binning. It should be noted that the fidelity of this continuous representation depends on the underlying data density, since the surface is approximated through interpolation or extrapolation in sparsely sampled regions.}

More importantly, this 3D representation enables a \textbf{robust global synthesis} through surface integration. By deriving a  global indicator (denoted as $\text{GMC}_g$) from the entire volume of the surface,  the resulting score is inherently distribution-agnostic, yielding a stable and ``true'' reflection of model capability across diverse application scenarios.

Extensive experiments on standard IQA datasets demonstrate that GMC enables more sophisticated performance analysis. The complementary strengths of different IQA models are systematically revealed, offering reliable evidence for model selection and integration. Moreover, GMC consistently outperforms traditional global metrics in both precision and reliability under varying quality distributions, highlighting its strong potential as a standardized evaluation measure.

\section{BACKGROUND}
\subsection{Image Quality Assessment Models}
\label{sec:related_work_iqa}
Objective IQA has evolved from early error-visibility measures to modern data-driven paradigms, paralleling our deepening understanding of the human visual system (HVS). Based on the availability of a reference image, these models are generally categorized into Full-Reference (FR-IQA) and No-Reference (NR-IQA) frameworks: (1)
\textbf{FR-IQA.} Early efforts focused on pixel-level fidelity, such as MSE and PSNR, which often fail to align with human perception due to oversimplified assumptions~\cite{mannos1974effects, wang2009mean, girod1993s}. A major paradigm shift occurred with the structural similarity (SSIM) index~\cite{wang2004image}, which prioritized structural fidelity over error visibility, leading to extensions like MS-SSIM~\cite{wang2003multiscale} and IW-SSIM~\cite{wang2010information}. Other classical approaches integrated information theory (VIF~\cite{sheikh2006image}) or gradient features (FSIM~\cite{zhang2011fsim}, GMSD~\cite{xue2013gradient}). Recently, perceptual comparisons have migrated to deep feature spaces, leveraging pre-trained networks (LPIPS~\cite{zhang2018unreasonable}, DISTS~\cite{ding2020image}) and advanced statistical distances (DeepWSD~\cite{liao2022deepwsd}, DeepDC~\cite{zhu2022deepdc}, DMM~\cite{chen2025debiased}) to capture complex HVS nonlinearities.
(2) \textbf{NR-IQA.} NR-IQA models aim to assess quality without a reference, initially relying on Natural Scene Statistics (NSS) across various domains, including spatial (NIQE~\cite{mittal2012making}), wavelet~\cite{moorthy2010two}, and DCT~\cite{saad2012blind}. Alternative hand-crafted methods explored the free-energy principle to estimate perceptual uncertainty~\cite{zhai2011psychovisual, gu2013no, chen2022no}. The field has since been dominated by deep learning architectures, ranging from early CNNs (CNNIQA~\cite{kang2014convolutional}, DBCNN~\cite{zhang2018blind}) to advanced Transformer-based and meta-learning models (HyperIQA~\cite{Su_2020_CVPR}, MetaIQA~\cite{zhu2020metaiqa}, MUSIQ~\cite{ke2021musiq}, LIQE~\cite{zhang2023blind}, TOPIQ~\cite{chen2024tip}). To enhance generalization, strategies like continual and transfer learning have been widely adopted~\cite{zhang2024task, chen2021learning}. Most recently, Large Multimodal Models (LMMs) have emerged as powerful zero-shot evaluators. Following foundational benchmarks like Q-Bench~\cite{wu2023qbench}, specialized LMM-based models such as Q-Align~\cite{wu2023qalign}, Dog-IQA~\cite{liu2024dog}, Compare2Score\cite{zhu2024adaptive}, Q-Debias\cite{chen2026mitigating}, and VisualQuality-R1~\cite{wu2025visualquality} have set new frontiers by aligning large-scale generative knowledge with human quality perception. 
Despite these advancements, the reliance on monolithic global metrics persists, leaving the local
 and fine-grained behavioral analysis of these diverse models is still largely unexplored.

\subsection{Performance Evaluation for IQA Models}
The rapid development of IQA models raises a fundamental question: how to reliably characterize model behavior and select appropriate methods for different application scenarios. Conventional evaluation protocols primarily rely on global correlation-based metrics, including the PLCC, SRCC, and KRCC, which compress model behavior into a single dataset-level statistic. To partially alleviate the uniform treatment of ranking errors, Wu \etal proposed Perceptual Weighted Rank Correlation (PWRC), which introduces perceptual weighting and ignores pairs with imperceptible quality differences\cite{wu2018perceptually}. However, PWRC remains a global metric and is therefore incapable of characterizing how performance varies across different levels of quality or quality differences.
Beyond correlation-based evaluation, several counter-example-driven protocols have been proposed to expose model failure cases through targeted stress testing. The Maximum Differentiation (MAD) competition~\cite{wang2008maximum} and its multi-model extension gMAD~\cite{ma2018group} formulate evaluation as a falsification process by searching for image pairs that maximally contradict a target model, with performance summarized in terms of aggressiveness and resistance. Complementarily, Ma \etal \cite{ma2016waterloo}introduced large-scale computational stress tests, including the D-test for pristine-versus-distorted discrimination, the L-test for monotonicity across distortion levels, and the P-test based on billions of automatically generated perceptually discriminable pairs (DIPs) inferred from agreement among multiple full-reference IQA models. These tests reveal extensive failure cases and provide fine-grained behavioral insights that are not captured by conventional benchmarks.
From a psychophysical perspective, Eigen-Distortions~\cite{berardino2017eigen} analyze extremal sensitivity directions of an IQA model via the Fisher Information Matrix and compare model predictions with human detection thresholds, yielding an absolute measure of perceptual alignment. While highly informative, this approach is computationally demanding and not intended for large-scale or routine model comparison.

\section{Method}
\subsection{Preliminary: Generalized Correlation Coefficient}

Let $x = \{x_i\}_{i=1}^n$ and $y = \{y_i\}_{i=1}^n$ be two sets of real-valued variables associated with $n$ items, such as predicted and ground-truth image quality scores, respectively. The \textit{Generalized Correlation Coefficient} (GCC), as proposed by Kendall and Gibbons\cite{kendall1990j}, is defined as:
\begin{equation}
\Gamma = \frac{\sum_{i,j=1}^{n} a_{ij} b_{ij}}{\sqrt{\sum_{i,j=1}^{n} a_{ij}^2} \cdot \sqrt{\sum_{i,j=1}^{n} b_{ij}^2}}.
\label{eq:gcc}
\end{equation}
Here, $a_{ij}$ and $b_{ij}$ are antisymmetric functions that quantify the relational difference between item pairs, \ie, $a_{ij} = -a_{ji}$ and $b_{ij} = -b_{ji}$.

\begin{figure*}[htbp] 
\centering \includegraphics[width=\textwidth]{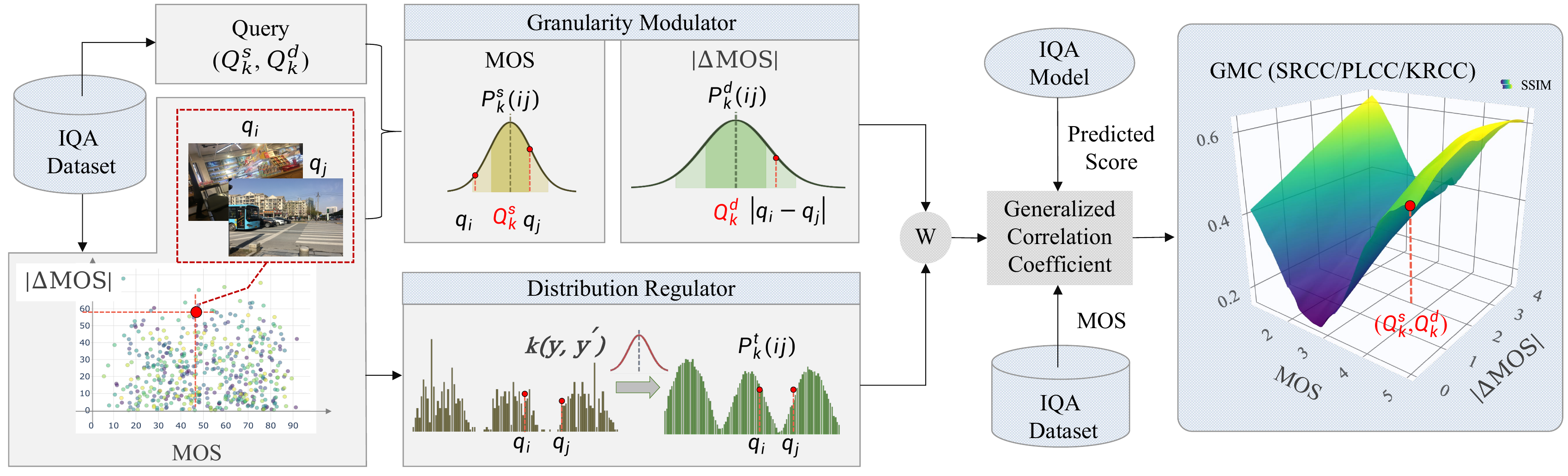} 
\caption{\rc{Overview of the proposed 3D GMC performance surface for fine-grained IQA evaluation. The framework consists of a \textbf{Granularity Modulator} for analyzing local performance variations conditioned on MOS and $|\Delta$MOS$|$, and a \textbf{Distribution Regulator} for mitigating biases from non-uniform quality distributions. The resulting correlation surface represents performance as a joint function of MOS and $|\Delta$MOS$|$.}}
\label{fig:flow} 
\end{figure*}

\subsubsection*{GCC Instantiations: PLCC, SRCC, and KRCC} By choosing appropriate definitions for $a_{ij}$ and $b_{ij}$, Eqn.~\eqref{eq:gcc} recovers several well-known correlation metrics\cite{wu2018perceptually}:
\begin{itemize}
\rc{
    \item \textbf{Pearson Linear Correlation Coefficient (PLCC):}
    
    Let $p_i = x_i$ (\ie, predicted scores), $q_i = y_i$ (\ie, MOS), and define
    \begin{equation}
    a_{ij} = p_i - p_j, \quad b_{ij} = q_i - q_j .
    \label{eq:plcc_def}
    \end{equation}
    Substituting Eqn.~\eqref{eq:plcc_def} into Eqn.~\eqref{eq:gcc}, it is straightforward to show that Eqn.~\eqref{eq:gcc} reduces to
    \begin{equation}
    \Gamma
    = \frac{\sum_i (p_i - \bar{p})(q_i - \bar{q})}
    {\sqrt{\sum_i (p_i - \bar{p})^2} \cdot \sqrt{\sum_i (q_i - \bar{q})^2}},
    \end{equation}
    where $\bar{p}$ and $\bar{q}$ denote mean quality. This is exactly the standard form of PLCC.\\}

% \begin{itemize}
%     \item \textbf{Pearson Linear Correlation Coefficient (PLCC):}
    
%     Let $p_i = x_i$ (\ie, predicted scores), $q_i = y_i$ (\ie, MOS), and define
%     \begin{equation}
%     a_{ij} = p_i - p_j, \quad b_{ij} = q_i - q_j .
%     \label{eq:plcc_def}
%     \end{equation}
%     Substituting Eqn.~\eqref{eq:plcc_def} into Eqn.~\eqref{eq:gcc}, the numerator becomes
%     \begin{equation}
%     \begin{aligned}
%     \sum_{i,j} (p_i - p_j)(q_i - q_j)
%     &= \sum_{i,j} \big(p_i q_i + p_j q_j - p_i q_j - p_j q_i \big) \\
%     &= 2n \sum_i (p_i - \bar{p})(q_i - \bar{q}) .
%     \end{aligned}
%     \end{equation}

%     and the denominator satisfies
%     \begin{equation}
%     \begin{aligned}
%     &\sqrt{\sum_{i,j} (p_i - p_j)^2}\,
%     \sqrt{\sum_{i,j} (q_i - q_j)^2}\\
%     &= \sqrt{2n \sum_i (p_i - \bar{p})^2}\,
%        \sqrt{2n \sum_i (q_i - \bar{q})^2} \\
%     &= 2n \sqrt{\sum_i (p_i - \bar{p})^2}\,
%          \sqrt{\sum_i (q_i - \bar{q})^2}.
%     \end{aligned}
%     \end{equation}

%     where $\bar{p}$ and $\bar{q}$ denote mean quality.
%     Therefore,  Eqn.~\eqref{eq:gcc} reduces to
%     \begin{equation}
%     \Gamma
%     = \frac{\sum_i (p_i - \bar{p})(q_i - \bar{q})}
%     {\sqrt{\sum_i (p_i - \bar{p})^2} \cdot \sqrt{\sum_i (q_i - \bar{q})^2}},
%     \end{equation}
%     which is exactly the standard form of PLCC.\\

    \item \textbf{Spearman Rank Correlation Coefficient (SRCC):}

    Let $p_i = \text{rank}(x_i)$, $q_i = \text{rank}(y_i)$, and again use Eqn.~\eqref{eq:plcc_def}. This yields the SRCC, which measures monotonic correlation via rank differences.\\

    \item \textbf{Kendall Rank Correlation Coefficient (KRCC):}
    Let $p_i = x_i$ and $q_i = y_i$, and define
    \begin{equation}
    a_{ij} = \operatorname{sgn}(p_i - p_j), \quad
    b_{ij} = \operatorname{sgn}(q_i - q_j),
    \label{eq:krcc_def}
    \end{equation}
    where the sign function is defined as
    \begin{equation}
    \operatorname{sgn}(z) =
    \begin{cases}
    1, & \text{if } z > 0,\\
    -1, & \text{if } z < 0,\\
    0, & \text{if } z = 0.
    \end{cases}
    \label{eq:sign}
    \end{equation}
    Then, Eqn.~\eqref{eq:gcc} reduces to the KRCC.

\end{itemize}
These formulations illustrate that traditional correlation metrics are specific instantiations of the GCC framework, with their behavior governed by the choice of $a_{ij}$ and $b_{ij}$.

\subsection{Limitations of Global Correlation-Based  Metrics}

Despite their widespread use, PLCC, SRCC, and KRCC exhibit several fundamental limitations when used to evaluate IQA models:

\begin{enumerate}
    \item \textbf{Lack of Fine-Grained Evaluability.} 
    All three metrics aggregate over all image pairs equally, implicitly assuming uniform perceptual relevance. Consequently, they fail to resolve how model performance varies across different absolute quality levels (MOS) or pairwise quality differences ($|\Delta$MOS$|$). As illustrated in Fig.~\ref{fig:abs}, CLIP-IQA achieves superior performance on high-MOS images, whereas NIQE is more discriminative when image pairs exhibit small $|\Delta$MOS$|$ values. Global correlation metrics, by collapsing performance into a single scalar, completely obscure these complementary behaviors.\\
    
    \item \textbf{Sensitivity to Quality Distribution.} 
    PLCC, SRCC, and KRCC are highly sensitive to the distribution of quality scores. As shown in Fig.~\ref{fig:dist}, two subsets sampled from the PIPAL dataset with identical sizes but differing MOS distributions produce reversed model rankings under PLCC and SRCC. This sensitivity arises from the uniform treatment of all $(i,j)$ pairs in the correlation computation, which conflates model capability with the specific characteristics of the test set. Ideally, an evaluation metric should remain robust to variations in the underlying MOS distribution, providing stable and consistent comparisons across datasets.\\

\item \textbf{Inability to Characterize Performance Landscape.} 
Global metrics fail to capture the underlying {landscape} of model behavior, whereas the geometric properties of a continuous correlation surface provide a formal measure of \textit{structural robustness}. For example, a fragmented landscape reveals latent instabilities and hypersensitivity to quality perturbations that are invisible to scalar summaries. Furthermore, the performance gradient across this surface serves as \textbf{optimization guidance}. By visualizing transition boundaries and perceptual constraints, such as sensitivity drops at specific bitrates, researchers can identify generalization limits, offering a systematic path for targeted IQA algorithmic refinement.
\end{enumerate}

\subsection{Our GMC Measure}
Given a test set with $n$ images,  predicted scores of the evaluated IQA model
\(\displaystyle P = \{p_1, p_2, \ldots, p_n\}\) 
and the corresponding ground-truth MOS values 
\(\displaystyle Q = \{q_1, q_2, \ldots, q_n\}\), 
our goal is to estimate the model IQA performance both \emph{locally} (at specific MOS and $|\Delta$MOS$|$) and \emph{robustly} (against imbalanced MOS distributions).  To achieve this, we model the IQA model performance visualization in a 3D manner (shown as Fig.~\ref{fig:flow}), spanning MOS (x-axis) and  ($|\Delta$MOS$|$), with the z-axis representing weighted correlation indicator (\eg, PLCC, SRCC, or KRCC). 
\rc{To motivate our design, consider a \textit{hypothetical} approach where one constructs local subsets via hard thresholding, \ie, selecting pairs satisfying $|q_i - Q^s_k| < \delta$ and $|q_i - q_j| < Q^d_k$ with a predefined threshold $\delta$. In this scheme, a point in the 3D space is represented by $(Q^s_k, Q^d_k, V_k)$, where $Q^s_k$ denotes the target MOS, $Q^d_k$ represents the $|\Delta$MOS$|$ of interest, and $V_k$ reflects model performance over the selected subset. By sampling different representatives, the performance landscape can, in principle, be obtained via plane fitting. However, this approach suffers from three critical limitations: selecting an appropriate $\delta$ is nontrivial; binary selection creates discontinuities between neighboring points; and subset construction cannot ensure uniform quality distributions. Consequently, minor variations induce large fluctuations in $V_k$, rendering plane fitting unreliable.}
%In general, a point in the 3D space can be represented by $(Q^s_k, Q^d_k, V_k)$, where $k$ is the sampled point index. Here, $Q^s_k$ denotes the target MOS, $Q^d_k$ represents the $|\Delta$MOS$|$ of interest, and $V_k$ reflects model performance over the subset of image pairs satisfying $|q_i - Q^s_k| < \delta$ and $|q_i - q_j| < Q^d_k$, with $\delta$ as a predefined threshold. By sampling different representatives $Q^s_k$ and $Q^d_k$, the performance landscape can, in principle, be obtained via plane fitting of the sampled points. However, selecting an appropriate $\delta$ is nontrivial, and the binary selection of pairs using $|q_i - Q^s_k| < \delta$ and $|q_i - q_j| < Q^d_k$ creates discontinuities between neighboring points. Consequently, minor variations in MOS or $\Delta$MOS can induce large fluctuations in $V_k$, rendering plane fitting unreliable.
%In addition, the subset construction at each sampled point still can not ensure the uniform quality distribution, leading to a biased performance measure. 
To account for this, build upon the GCC model (Eqn.~(\ref{eq:gcc})), we respectively propose a  \emph{Granularity  Modulator} and \emph{Distribution  Regulator} to achieve a smooth and robust performance estimation at each fine-grained point as follows,
\begin{equation}
\Gamma_{k} = \frac{\sum_{i,j=1}^{n}  w_k^{ij}a_{ij} b_{ij}}{\sqrt{\sum_{i,j=1}^{n} w_k^{ij} a_{ij}^2} \cdot \sqrt{\sum_{i,j=1}^{n}w_k^{ij} b_{ij}^2}},
\label{eq:gcck}
\end{equation}
where
\begin{equation}
w_k^{ij} =
\underbrace{
    \underset{\text{Absolute Quality Score}}{\underline{P_k^{s}(i, j)}} \times 
        \underset{\text{Pairwise Quality Difference}}{\underline{P_k^{d}(i, j)}}
}_{\text{Granularity  Modulator}} \times
\underbrace{\underset{\text{Quality Distribution}}{\underline{P_k^{t}(i, j)}}}_{\text{Distribution  Regulator}},
\label{eqn:pk}
\end{equation}
and  $P_k^{s}(i, j)$ and $P_k^{d}(i, j)$ are the contribution weights of $i$-th and $j$-th images for the correlation measure at the specific quality scale $Q^s_k$  and quality difference  $Q^d_k$  in our Granularity  Modulator module and $P_k^{t}(i, j)$ is distribution regularization term in our Distribution  Regulator module. When we respectively adopt the PLCC, SRCC and KRCC as the localized performance indicator, the Eqn.~(\ref{eq:gcck}) can be formed by:
{\small
\begin{align}
\Gamma_{k}(\text{PLCC}) &= 
\frac{
    \sum_{i,j=1}^{n} w_k^{ij} (p_i - p_j)(q_i - q_j)
}{
    \sqrt{\sum_{i,j=1}^{n} w_k^{ij} (p_i - p_j)^2}
    \sqrt{\sum_{i,j=1}^{n} w_k^{ij} (q_i - q_j)^2}
},
\label{eq:GMC_plcc}
\end{align}}
{\small
\begin{align}
\Gamma_{k}(\text{SRCC}) &= 
\frac{
    \sum_{i,j=1}^{n} w_k^{ij} (r_{p_i} - r_{p_j})(r_{q_i} - r_{q_j})
}{
    \sqrt{\sum_{i,j=1}^{n} w_k^{ij} (r_{p_i} - r_{p_j})^2}
    \sqrt{\sum_{i,j=1}^{n} w_k^{ij} (r_{q_i} - r_{q_j})^2}
},
\label{eq:GMC_srcc}
\end{align}
}

{\footnotesize
\begin{align}
\Gamma_{k}(\text{KRCC}) &= 
\frac{
    \sum_{i,j=1}^{n} w_k^{ij} \, \text{sgn}(p_i - p_j)\, \text{sgn}(q_i - q_j)
}{
    \sqrt{\sum_{i,j=1}^{n} w_k^{ij} \, \text{sgn}(p_i - p_j)^2}
    \sqrt{\sum_{i,j=1}^{n} w_k^{ij} \, \text{sgn}(q_i - q_j)^2}
},
\label{eq:GMC_krcc}
\end{align}}
where $r_{p_i}$ and $r_{q_i}$ are the rank of ${p_i}$ and ${p_j}$, respectively. The details of the two modules are described in the following.

\subsubsection{{Granularity  Modulator}}
At a query MOS $Q^s_k$, we adopt a Gaussian model to estimate the contribution weight of each image pair ($I_i$ and $I_j$) in the mode performance measure smoothly. In particular,
\begin{equation}
P_k^{s}(i, j) = \exp\left(-\frac{(Q^s_k-q_i)^2}{2\sigma_i^2}  -\frac{(Q^s_k-q_j)^2}{2\sigma_j^2} \right) ,
\end{equation}
where \( \sigma_i \) and \( \sigma_j \) are the standard deviations of the subjective quality ratings for images $I_i$ and $I_j$. Herein, the standard deviations can be obtained from the test dataset if it provides or estimated by a Beta distribution followed by\cite{ferrari2004beta}.
Herein, the reason we adopt the  Gaussian modeling lies in that human ratings (denoted as $R_i$ and $R_j$ for $I_i$ and $I_j$) usually follows a Gaussian distribution with the MOS,~$q_i$ or $q_j$ as the mean value\cite{zhang2021uncertainty,you2024descriptive}, \ie, $R_i \sim \mathcal{N}\left(q_i, \sigma_i^2\right), R_j \sim \mathcal{N}\left(q_j, \sigma_j^2\right)$. \rc{ Assuming that the quality scores $q_i$ and $q_j$ are statistically independent, the joint probability of both falling at the target MOS $Q_k^s$ is exactly equal to the product of their individual Gaussian probabilities:
\begin{equation}
    P(q_i=Q_k^s \wedge q_j=Q_k^s) = P(q_i=Q_k^s) \cdot P(q_j=Q_k^s).
\end{equation}
}

\rc{Analogously, let $R_i$ and $R_j$ denote the underlying random variables corresponding to the subjective ratings of images $I_i$ and $I_j$, respectively. Under the one-shot estimation assumption, the MOS realizations $q_i$ and $q_j$ are treated as estimates of their corresponding means. Consequently, the signed quality difference
$
R_i-R_j
$
follows a Gaussian distribution:
\begin{equation}
R_i-R_j
\sim
\mathcal{N}
\left(
q_i-q_j,\,
\sigma_i^2+\sigma_j^2
\right).
\end{equation}
\textit{
Since our GMC formulation aims to characterize the discrimination capability of IQA models under different perceptual difference magnitudes, we adopt the absolute quality difference irrespective of direction. Accordingly, the corresponding random variable
$
|R_i-R_j|
$
follows a Folded Normal distribution. Therefore, the probability density that the image pair $(I_i,I_j)$ falls at the target absolute quality difference $Q_k^d$ is evaluated as:
}
\begin{equation}
\begin{aligned}
P_k^d(i,j)
&=
\frac{1}
{\sqrt{2\pi(\sigma_i^2+\sigma_j^2)}}
\Bigg[
\exp
\left(
-\frac{
(Q_k^d-(q_i-q_j))^2
}{
2(\sigma_i^2+\sigma_j^2)
}
\right)
\\
&\quad +
\exp
\left(
-\frac{
(Q_k^d+(q_i-q_j))^2
}{
2(\sigma_i^2+\sigma_j^2)
}
\right)
\Bigg],
\quad Q_k^d \ge 0.
\end{aligned}
\end{equation}
}
% Assuming that the quality scores \( q_i \) and \( q_j \) are statistically independent, the joint probability of both falling at the target MOS \( Q_k^s \) can be approximated as the product of their individual Gaussian probabilities:
% \begin{equation}
% P(q_i = Q_k^s \land q_j = Q_k^s) \approx P(q_i = Q_k^s) \cdot P(q_j = Q_k^s).
% \end{equation}
% Analogously,  the quality difference $|q_i -q_j|$ also follows a Gaussian distribution, $\mathcal{N}\left(q_i-q_j, \sigma_i^2+\sigma_j^2\right)$. The probability that the image pair \( (I_i, I_j) \) falls within the target quality difference \( Q^d_k \) can be obtained by:
% \begin{equation}
% P_k^{d}(i, j) = \exp\left(-\frac{(Q^d_k - |q_i - q_j|)^2}{\sigma_i^2 + \sigma_j^2}\right).
% \end{equation}
Based on the weight modulation results $P_k^{s}(i, j)$ and $P_k^{d}(i, j)$, our Granularity  Modulator acts as a localized observer, enabling a fine-grained assessment of the model capability on each level of MOS and $|\Delta\text{MOS}|$.

\subsubsection{{Distribution  Regulator}}
\begin{figure}[!t] 
\centering 
\includegraphics[width=0.5\textwidth]{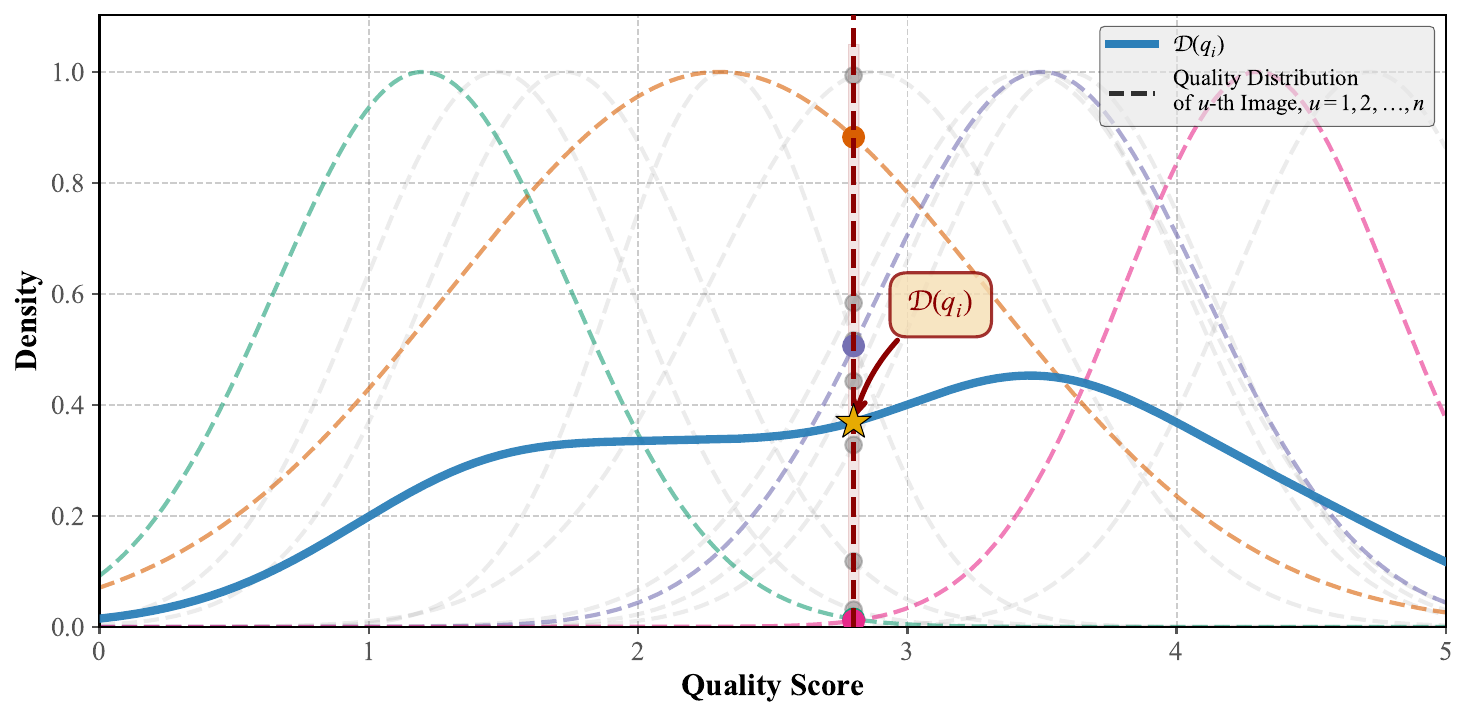} 
\caption{Density estimation at a target quality score $q_i$.
Each sample $q_u$ contributes to the density at $q_i$ via a Gaussian kernel with sample-dependent standard deviation.}
\label{fig:dense} 
\end{figure}

To mitigate the correlation bias caused by the imbalanced MOS distribution,  we  propose a \textit{Distribution  Regulator} in our GMC.   Specifically,  we define the $P_k^{t}(i, j)$ in Eqn.~(\ref{eqn:pk}) by:
\begin{equation}
P_k^{t}(i, j) = \frac{1}{\mathcal{D}(q_i)} \cdot \frac{1}{\mathcal{D}(q_j)},
\label{eq:weight}
\end{equation}
where $\mathcal{D}(q_i)$ and $\mathcal{D}(q_j)$ are the estimated density of quality $q_i$ and $q_j$. Herein, take $\mathcal{D}(q_i)$ as an example, we considered two cases in the density estimation:
\begin{enumerate}
    \item \textbf{Standard deviation is available.}  
    As illustrated in Fig.~\ref{fig:dense}, when image-wise standard deviations of quality ratings are available in the test set, we estimate the local density at $q_i$ by accumulating Gaussian kernels centered at all samples:
    \begin{equation}
    \mathcal{D}(q_i) = \frac{1}{n} \sum_{u=1}^{n} \exp\!\left(-\frac{(q_u - q_i)^2}{2\sigma_u^2}\right),
    \end{equation}
    where $\sigma_u$ denotes the sample-specific standard deviation associated with $q_u$.\\

    \item \textbf{Standard deviation is unavailable.} In this case, we introduce a symmetric kernel to extract a kernel-smoothed density estimation. Specifically, we first discretize the normalized MOS range (\eg, $[1, 100]$) into $\mathcal{Y}$ equal-width bins and compute the sample frequency of each bin, then the $ \mathcal{D}(q_i)$ can be estimated by,
    
    \begin{equation}
    \mathcal{D}(q_i) \triangleq \int_{\mathcal{Y}} \mathrm{k}\left(y, y^{\prime}\right) p(y) dy,
     \end{equation}  
     \rc{where $p(y)$ is the sample frequency of the bin in which $q_i$ is located. $k(\cdot,\cdot)$ is a symmetric and translation-invariant smoothing kernel. Formally, for continuous variables $z$ and $z'$, the kernel satisfies $k(z,z') = k(z',z)$ and $\nabla_z k(z,z') + \nabla_{z'} k(z',z) = 0$ \cite{yang2021delving}. By evaluating this kernel at the discrete bin values $y$ and $y'$, we convolve it with the empirical density distribution to extract a kernel-smoothed version that accounts for the overlap in information of data samples in nearby quality bins. Without loss of generality, we adopt the Gaussian kernel in our GMC measure.}\\
    % where $p(y) = p(\lfloor q_i \rfloor) $ is the sample frequency of the bin that $q_i$ located. $\mathrm{k}(\cdot,\cdot)$ is a symmetric kernel satisfy $\mathrm{k}\left(y, y^{\prime}\right)=\mathrm{k}\left(y^{\prime}, y\right)$ and $\nabla_y \mathrm{k}\left(y, y^{\prime}\right)+\nabla_{y^{\prime}} \mathrm{k}\left(y^{\prime}, y\right)=0$, $\forall y, y^{\prime} \in \mathcal{Y}$ \cite{yang2021delving}. Without loss of generality, we adopt the Gaussian kernel in our GMC measure.\\
\end{enumerate}

\noindent \textbf{Design Rationale: Why Kernel Smoothing Is Preferred in Quality Density Estimation?}
Though the sample frequency of each bin is widely adopted for the category density estimation in the classification task, we argue that the two limitations lie:  (1) The hard assignment of scores to bins, ignoring uncertainty in subjective ratings (\eg, an image with MOS $70.5$ might reasonably contribute to both the $[66$–$70]$ and $[71$–$75]$ bins); and (2) Certain target values may have no data at all, and the empty bins can cause division-by-zero errors in Eqn.~(\ref{eq:weight}),  motivating our interpolation manner density smoothing. In Sec.~\ref {sec:as}, our experiments also verify that our kernel-smoothed density estimation strategy results in more robust IQA model performance evaluation.

\subsubsection{3D Performance Surface Modeling}
Based upon Eqn.~(\ref{eq:gcck}), we model the IQA model performance in a 3D visualization space ($Q^s_k$, $Q^d_k$, $V_k$) which captures not only the perceptual quality but also integrates the concept of quality discrimination, allowing us to jointly assess model performance in terms of both quality estimation and quality differentiation. To better compare the performance of different IQA models and analyze the performance topology, the performance plane can be fitted when different points ($k$) are sampled. Herein, the reliability of the plane fitting is higher as the number of sampled points increases.  However, exhaustively traversing all possible quality scales and difference across the dataset is  inefficient, especially when the dataset volume is large. \\

\noindent \textbf{Latin Hypercube Sampling.} 
\rc{To efficiently explore the parameter space and ensure a representative selection of samples, we employ Latin Hypercube Sampling (LHS) \cite{mckay2000comparison, santner2003design}\footnote{\rc{For efficient implementation, we assume a uniform marginal distribution along each dimension, avoiding complicated Cumulative Distribution Function (CDF) estimation.}}}
% To ensure uniform and efficient sampling within the joint spatial space defined by the horizontal and vertical offsets \(( Q^s,  Q^d)\), we employ the 2D \textit{Latin Hypercube Sampling} (LHS) 
strategy to generate a representative set of \( K \) sampling points \( \{(Q^s_k, Q^d_k)\}_{k=1}^K \). Specifically, LHS partitions the value range of each variable, \ie, \([ Q^s_{\min},  Q^s_{\max}]\) and \([ Q^d_{\min},  Q^d_{\max}]\) into \( K \) non-overlapping intervals of equal width. For each sampling index \( k \), a random value is drawn from a unique interval along each dimension, ensuring stratified coverage without redundancy. The sampling coordinates are computed as:
\begin{align}
Q^s_k &= \frac{\pi_x(k) - u_k}{K} \cdot ( Q^s_{\max} -  Q^s_{\min}) +  Q^s_{\min},\\
Q^d_k &= \frac{\pi_y(k) - u_k}{K} \cdot ( Q^d_{\max} -  Q^d_{\min}) +  Q^d_{\min},
\label{eq:lhs}
\end{align}
where \( \pi_x \) and \( \pi_y \) denote independent random permutations of \(\{1, 2, \dots, K\}\), and \( u_k \sim \mathcal{U}(0,1) \) is a uniform random variable used to perturb sampling within each interval. This formulation guarantees that each projection of the \( K \) samples covers the domain uniformly, yielding improved statistical efficiency and coverage compared to naïve random sampling.\\

\noindent \textbf{Performance Surface Modeling.} Once the weighted correlation score \(\Gamma_{k}\) is computed for each sampling point, we fit a continuous 3D performance response surface to characterize the model’s behavior across the joint space of quality scale and quality difference. This fitting process can be expressed as:

\begin{equation}
\text{GMC}(Q^s_k, Q^d_k) = \hat{\Gamma}(Q^s_k, Q^d_k) \approx \mathcal{F}\left(\{(Q^s_k, Q^d_k)\}_{k=1}^N\right),
\label{eq:fit}
\end{equation}
where \(\mathcal{F}(\cdot)\) denotes a two-dimensional nonparametric function approximation based on Local Linear Kernel Regression. This method performs weighted linear modeling within local neighborhoods, effectively reducing boundary bias. \\

\noindent \textbf{Global Performance Aggregation.}  In addition to the localized analysis, a global performance score can also be derived from our GMC (denoted as $\text{GMC}_g$) via a surface integration:

\begin{equation}
    \text{GMC}_g = \frac{1}{A} \int_{  Q^s_{\min}}^{  Q^s_{\max}} \int_{ Q^d_{\min}}^{ Q^d_{\max}} \hat{\Gamma}(x, y)\, dx \, dy,
\end{equation}
where \(A = ( Q^s_{\max} -  Q^s_{\min}) \times ( Q^d_{\max} -  Q^d_{\min})\) represents the normalization area formed by the ranges of $Q^s$ and $Q^d$. Compared with the traditional global indicators (\eg, PLCC, SRCC or KRCC), our derived indicator $\text{GMC}_g$  not only reflects the model's predictive capability for absolute image quality scales but also captures its sensitivity to quality difference discrimination with the distribution bias mitigated, offering a comprehensive and robust performance assessment of IQA model.

\begin{figure*}[t] 
\centering 
\includegraphics[width=\textwidth]{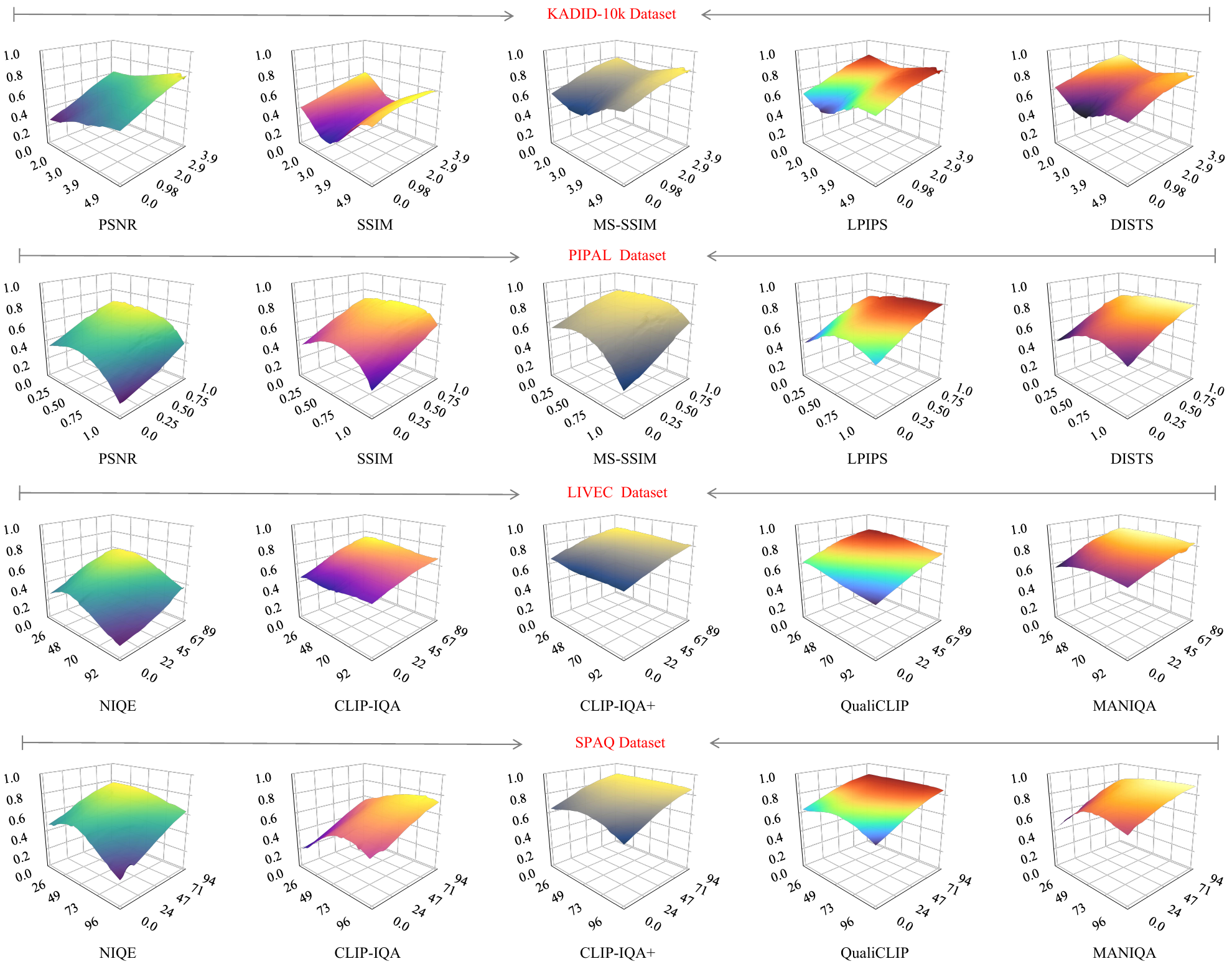} 
\caption{Visualization of correlation surfaces generated by different IQA models. The ``$\text{GMC}$" response (vertical axis) is shown as a function of the ``MOS" (left horizontal axis) and ``$|\Delta\text{MOS}|$" (right horizontal axis).
} 
\label{fig:3dvis} 
\end{figure*}

\begin{figure*}[t] 
\centering 
\includegraphics[width=\textwidth]{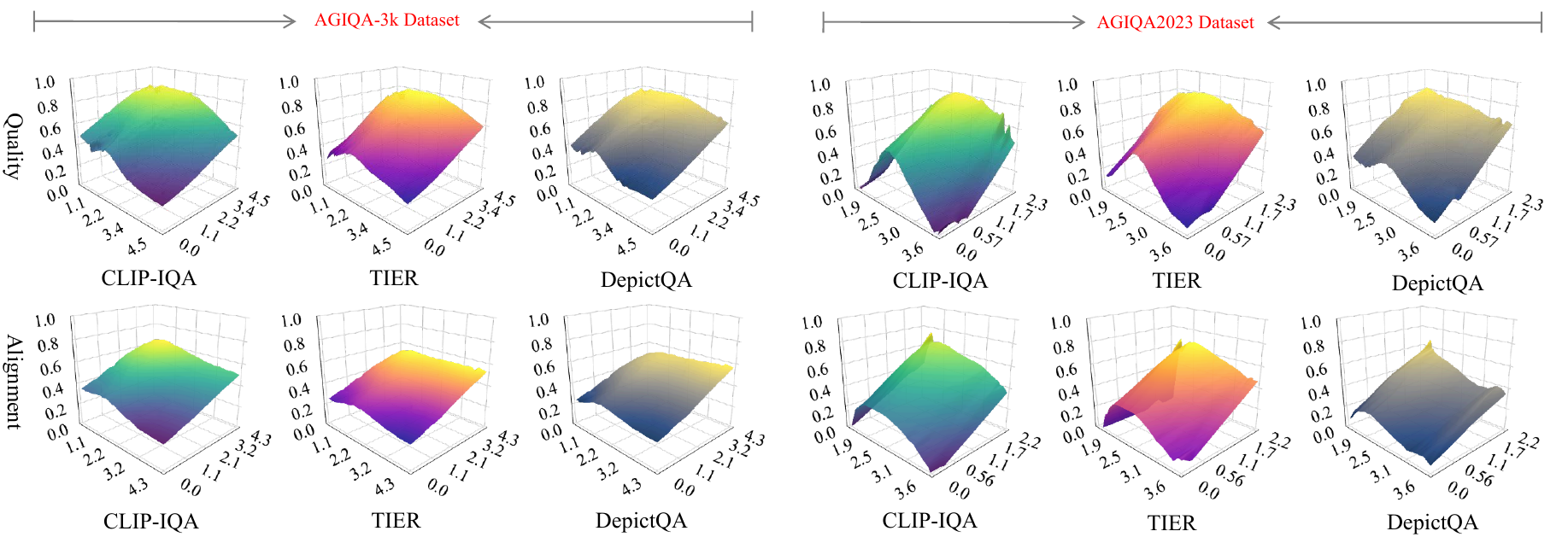} 
\caption{\rc{Visualization of correlation surfaces generated by different IQA models. The ``$\text{GMC}$" response (vertical axis) is shown as a function of the ``MOS" (left horizontal axis) and ``$|\Delta\text{MOS}|$" (right horizontal axis).
} }
\label{fig:3dvis-aigc} 
\end{figure*}

\section{Experiment}
\subsection{Experimental Settings}

We evaluate the proposed GMC measure on both FR and NR IQA models. For {FR-IQA}, five representative metrics are considered: PSNR, SSIM\cite{wang2004image}, MS-SSIM\cite{wang2003multiscale}, LPIPS \cite{zhang2018unreasonable}, and DISTS \cite{ding2020image}. These FR metrics are evaluated on the full KADID-10k database \cite{2019kadid} and the PIPAL dataset \cite{jinjin2020pipal}. 
For {NR-IQA}, we select five recent blind-quality models: NIQE\cite{Mittal2013SPL}, CLIP-IQA (CLIP-IQA and CLIP-IQA+)\cite{Wang2023AAAI}, {QualiCLIP}\cite{agnolucci2024qualityaware} and
{MANIQA}\cite{yang2022maniqa}. Each NR model is trained on the KonIQ-10k dataset \cite{Hosu2020KonIQ} and then tested on two large-scale ``in-the-wild" datasets: LIVE-Challenge (denoted as LIVEC)\cite{ghadiyaram2015massive} and SPAQ\cite{fang2020cvpr}. 
\rc{In addition, we extend our experiments to two AIGC benchmark datasets, AGIQA-3K \cite{li2023agiqa} and AIGCIQA2023 \cite{wang2023aigciqa2023}. We select three representative recent models, namely {CLIP-IQA} \cite{wang2023exploring}, {TIER} \cite{yuan2024tier}, and {DepictQA} \cite{you2024descriptive}, and conduct cross-dataset testing on two quality dimensions, \ie, \textit{Quality} and \textit{Alignment}.}
We discretize quality scores into $|\mathcal{Y}|=100$ bins and employ Latin hypercube sampling with $K=100$ sample points when computing GMC.  We adopt the SRCC, \ie, $\Gamma_{k}(\text{SRCC})$ in Eqn.~(\ref{eq:GMC_srcc}) for the localized performance evaluation by default.

\begin{table*}[!t]
\centering
\caption{\rc{Performance comparison of FR- and NR-IQA models across different IQA datasets in terms of PLCC, SRCC, $\text{GMC}_s$, and $\text{GMC}_d$.}}
\label{tab:gmc}
\scriptsize
\renewcommand{\arraystretch}{1.08} 
\setlength{\tabcolsep}{4pt} 
\newcolumntype{M}[1]{>{\centering\arraybackslash}m{#1}}

%%%%%%%%%%%%%%%%%%%%%%%%%%%% KADID %%%%%%%%%%%%%%%%%%%%%%%%%%%%
\resizebox{0.9\textwidth}{!}{
\begin{tabular}{ L{2.2cm} | M{1.05cm} M{1.05cm} M{1.05cm} | M{1.05cm} M{1.05cm} M{1.05cm} | M{1.05cm} M{1.05cm} M{1.05cm} M{1.05cm} }
\toprule

\rowcolor{lightcyan}
\multicolumn{11}{c}{\textbf{KADID-10k\cite{2019kadid} (FR)}} \\
\hline
\multirow{2}{*}{\textbf{Model}}  &
\multicolumn{3}{c|}{GMC$_s$} & \multicolumn{3}{c|}{GMC$_d$} & \multirow{2}{*}{PLCC} & \multirow{2}{*}{SRCC} & \multirow{2}{*}{KRCC}  & \multirow{2}{*}{PWRC} \\
\cline{2-4}\cline{5-7}
 & LQ & MQ & HQ & LD & MD & HD & & & \\
\hline
\textit{PSNR}      & 0.4509 & 0.5195 & 0.6113 & 0.4047 & 0.5204 & 0.6458 & 0.5557 & 0.6757 & 0.4876 & 21.6874 \\
\textit{SSIM}\cite{wang2004image}     & 0.3927 & 0.3286 & 0.5555 & 0.3443 & 0.4245 & 0.4918 & 0.5755 & 0.6188 & 0.4468 & 19.2923\\
\textit{MS-SSIM}\cite{wang2003multiscale}  & \textit{\gr{0.6448}} & \textit{\gr{0.6269}} & \textit{\gr{0.7357}} & \textit{\gr{0.5608}} & \textit{\gr{0.6729}} & \textit{\gr{0.7652}} & 0.6802 & \textbf{\br{0.8256}} & \textbf{\br{0.6350}} & \textit{\gr{25.9811}} \\
\textit{LPIPS}\cite{zhang2018unreasonable}    & 0.6363 & \textbf{\br{0.6453}} & \textbf{\br{0.7428}} & \textbf{\br{0.5590}} & \textbf{\br{0.6779}} & \textbf{\br{0.7790}} & \textit{\gr{0.7484}} & \textit{\gr{0.8224}} & \textit{\gr{0.6303}} & \textbf{\br{26.2150}} \\
\textit{DISTS}\cite{zhang2018unreasonable}    & \textbf{\br{0.6682}} & 0.6041 & 0.6867 & 0.5458 & 0.6586 & 0.7505 & \textbf{\br{0.8057}} & 0.8137 & 0.6254 & 25.4562 \\
\hline
\end{tabular}}

\vspace{4pt}

%%%%%%%%%%%%%%%%%%%%%%%%%%%% PIPAL %%%%%%%%%%%%%%%%%%%%%%%%%%%%
\resizebox{0.9\textwidth}{!}{
\begin{tabular}{ L{2.2cm} | M{1.05cm} M{1.05cm} M{1.05cm} | M{1.05cm} M{1.05cm} M{1.05cm} | M{1.05cm} M{1.05cm} M{1.05cm} M{1.05cm} }

\rowcolor{lightcyan}
\multicolumn{11}{c}{\textbf{PIPAL \cite{jinjin2020pipal} (FR)}} \\
\hline
\multirow{2}{*}{\textbf{Model}}  &
\multicolumn{3}{c|}{GMC$_s$} & \multicolumn{3}{c|}{GMC$_d$} & \multirow{2}{*}{PLCC} & \multirow{2}{*}{SRCC} & \multirow{2}{*}{KRCC} & \multirow{2}{*}{PWRC}\\
\cline{2-4}\cline{5-7}
 & LQ & MQ & HQ & LD & MD & HD & & & \\
\hline
\textit{PSNR}      & 0.5364 & 0.4842 & 0.3927 & 0.3769 & 0.4762 & 0.5860 & 0.4033 & 0.4065 & 0.2762 & 5.5463\\
\textit{SSIM}\cite{wang2004image}     & 0.5926 & 0.5853 & 0.5520 & 0.4665 & 0.5763 & 0.6893 & 0.4992 & 0.5041 & 0.3489 & 7.2886\\
\textit{MS-SSIM}\cite{wang2003multiscale}  & \textbf{\br{0.7049}} & 0.6406 & 0.5637 & \textbf{\br{0.5428}} & 0.6480 & 0.7525 & 0.5624 & 0.5621 & 0.3975 & 8.2772\\
\textit{LPIPS}\cite{zhang2018unreasonable}    & 0.6351 & \textbf{\br{0.6531}} & \textbf{\br{0.6808}} & 0.5419 & \textit{\gr{0.6556}} & \textit{\gr{0.7624}} & \textbf{\br{0.5827}} & \textbf{\br{0.5854}} & \textbf{\br{0.4099}} & \textbf{\br{8.7563}} \\
\textit{DISTS}\cite{zhang2018unreasonable}    & \textit{\gr{0.6414}} & \textit{\gr{0.6500}} & \textit{\gr{0.6733}} & \textit{\gr{0.5421}} & \textbf{\br{0.6560}} & \textbf{\br{0.7629}} & \textit{\gr{0.5797}} & \textit{\gr{0.5790}} & \textit{\gr{0.4069}} & \textit{\gr{8.6281}} \\
\hline
\end{tabular}
}
\vspace{4pt}

%%%%%%%%%%%%%%%%%%%%%%%%%%%% LIVEC %%%%%%%%%%%%%%%%%%%%%%%%%%%%
\resizebox{0.9\textwidth}{!}{
\begin{tabular}{ L{2.2cm} | M{1.05cm} M{1.05cm} M{1.05cm} | M{1.05cm} M{1.05cm} M{1.05cm} | M{1.05cm} M{1.05cm} M{1.05cm} M{1.05cm} }

\rowcolor{lightcyan}
\multicolumn{11}{c}{\textbf{LIVEC \cite{ghadiyaram2015massive} (NR)}} \\
\hline
\multirow{2}{*}{\textbf{Model}}  &
\multicolumn{3}{c|}{GMC$_s$} & \multicolumn{3}{c|}{GMC$_d$} & \multirow{2}{*}{PLCC} & \multirow{2}{*}{SRCC} & \multirow{2}{*}{KRCC} & \multirow{2}{*}{PWRC} \\
\cline{2-4}\cline{5-7}
 & LQ & MQ & HQ & LD & MD & HD & & & \\
\hline
\textit{NIQE}\cite{Mittal2013SPL}       & 0.4768 & 0.3741 & 0.2523 & 0.3035 & 0.4031 & 0.4965 & 0.4791 & 0.4495 & 0.3063 & 6.9598\\
\textit{CLIP-IQA}\cite{Wang2023AAAI}   & 0.6123 & 0.5854 & 0.5897 & 0.5152 & 0.6221 & 0.7005 & 0.6883 & 0.6955 & 0.5065 & 10.9756\\
\textit{CLIP-IQA+}\cite{Wang2023AAAI}  & \textit{\gr{0.7561}} & \textit{\gr{0.7133}} & \textit{\gr{0.7091}} & \textit{\gr{0.6606}} & \textit{\gr{0.7613}} & \textit{\gr{0.8254}} & \textit{\gr{0.8312}} & \textit{\gr{0.8045}} & \textit{\gr{0.6109}} & \textit{\gr{12.6591}} \\
\textit{QualiCLIP}\cite{agnolucci2024qualityaware}  & 0.7285 & 0.6540 & 0.6181 & 0.6013 & 0.7075 & 0.7799 & 0.7967 & 0.7553 & 0.5618 & 12.0622\\
\textit{MANIQA}\cite{yang2022maniqa}     & \textbf{\br{0.7640}} & \textbf{\br{0.7601}} & \textbf{\br{0.7508}} & \textbf{\br{0.6879}} & \textbf{\br{0.7892}} & \textbf{\br{0.8477}} & \textbf{\br{0.8401}} & \textbf{\br{0.8328}} & \textbf{\br{0.6404}} & \textbf{\br{13.0035}}\\
\hline
\end{tabular}
}
\vspace{4pt}

%%%%%%%%%%%%%%%%%%%%%%%%%%%% SPAQ %%%%%%%%%%%%%%%%%%%%%%%%%%%%
\resizebox{0.9\textwidth}{!}{
\begin{tabular}{ L{2.2cm} | M{1.05cm} M{1.05cm} M{1.05cm} | M{1.05cm} M{1.05cm} M{1.05cm} | M{1.05cm} M{1.05cm} M{1.05cm}  M{1.05cm}}

\rowcolor{lightcyan}
\multicolumn{11}{c}{\textbf{SPAQ \cite{fang2020cvpr} (NR)}} \\
\hline
\multirow{2}{*}{\textbf{Model}}  &
\multicolumn{3}{c|}{GMC$_s$} & \multicolumn{3}{c|}{GMC$_d$} & \multirow{2}{*}{PLCC} & \multirow{2}{*}{SRCC} & \multirow{2}{*}{KRCC} & \multirow{2}{*}{PWRC}\\
\cline{2-4}\cline{5-7}
 & LQ & MQ & HQ & LD & MD & HD & & & \\
\hline
\textit{NIQE}\cite{Mittal2013SPL}       & 0.6532 & 0.6296 & 0.5103 & 0.4977 & 0.6186 & 0.7119 & 0.6692 & 0.6928 & 0.4928 & 16.5133\\
\textit{CLIP-IQA}\cite{Wang2023AAAI}   & 0.4541 & 0.5968 & 0.6258 & 0.4548 & 0.5704 & 0.6668 & 0.6432 & 0.6653 & 0.4645 & 16.3178 \\
\textit{CLIP-IQA+}\cite{Wang2023AAAI}  & \textbf{\br{0.7882}} & \textit{\gr{0.7917}} & \textit{\gr{0.7685}} & \textbf{\br{0.7086}} & \textbf{\br{0.8137}} & \textbf{\br{0.8724}} & 0.8492 & \textbf{\br{0.8516}} & \textbf{\br{0.6492}} & \textit{\gr{20.7123}}\\
\textit{QualiCLIP}\cite{agnolucci2024qualityaware}  & \textit{\gr{0.7766}} & 0.7822 & 0.7613 & \textit{\gr{0.6970}} & \textit{\gr{0.8038}} & \textit{\gr{0.8649}} & \textit{\gr{0.8518}} & 0.8440 & 0.6386 & 20.5147\\
\textit{MANIQA}\cite{yang2022maniqa}     & 0.7100 & \textbf{\br{0.7969}} & \textbf{\br{0.8090}} & 0.6901 & 0.7983 & 0.8629 & \textbf{\br{0.8552}} & \textit{\gr{0.8509}} & \textit{\gr{0.6471}} & \textbf{\br{20.8074}}\\
\bottomrule
\end{tabular}}
\end{table*}

\begin{table*}[t]
\centering
\caption{\rc{Performance comparison of AGIQA models across AIGC datasets in terms of PLCC, SRCC, KRCC, PWRC, $\text{GMC}_s$, and $\text{GMC}_d$.}}
\label{tab:gmc-aigc}
\scriptsize
\renewcommand{\arraystretch}{1.08} 
\setlength{\tabcolsep}{4pt} 
\newcolumntype{M}[1]{>{\centering\arraybackslash}m{#1}}

%%%%%%%%%%%%%%%%%%%%%%%%%%%% AGIQA-3K %%%%%%%%%%%%%%%%%%%%%%%%%%%%
\resizebox{0.9\textwidth}{!}{
\begin{tabular}{ L{2.2cm} | M{1.05cm} M{1.05cm} M{1.05cm} | M{1.05cm} M{1.05cm} M{1.05cm} | M{1.05cm} M{1.05cm} M{1.05cm} M{1.05cm} }
\toprule

\rowcolor{lightcyan}
\multicolumn{11}{c}{\textbf{AGIQA-3K (Quality)}} \\
\hline

\multirow{2}{*}{\textbf{Model}}  &
\multicolumn{3}{c|}{GMC$_s$} & \multicolumn{3}{c|}{GMC$_d$} &
\multirow{2}{*}{PLCC} & \multirow{2}{*}{SRCC} & \multirow{2}{*}{KRCC} & \multirow{2}{*}{PWRC} \\
\cline{2-4}\cline{5-7}

 & LQ & MQ & HQ & LD & MD & HD & & & & \\
\hline

\textit{CLIP-IQA} \cite{wang2023exploring}  
& \textbf{\br{0.6401}} & \textit{\gr{0.5243}} & 0.4080
& \textbf{\br{0.3901}} & \textbf{\br{0.5738}} & 0.7068
& 0.6645 & 0.6137 & 0.4278 & 13.6306 \\

\textit{TIER}  \cite{yuan2024tier}
& 0.5958 & \textbf{\br{0.5542}} & \textbf{\br{0.4610}}
& \textit{\gr{0.3888}} & \textit{\gr{0.5736}} & \textbf{\br{0.7084}}
& \textbf{\br{0.6717}} & \textbf{\br{0.6525}} & \textbf{\br{0.4594}} & \textbf{\br{14.2661}} \\

\textit{DepictQA}  \cite{you2024descriptive}
& \textit{\gr{0.6120}} & 0.5201 & \textit{\gr{0.4514}}
& 0.3873 & 0.5726 & \textit{\gr{0.7071}}
& \textit{\gr{0.6646}} & \textit{\gr{0.6184}} & \textit{\gr{0.4494}} & \textit{\gr{13.7831}} \\

\hline
\end{tabular}}

\vspace{4pt}

%%%%%%%%%%%%%%%%%%%%%%%%%%%% AGIQA-3K %%%%%%%%%%%%%%%%%%%%%%%%%%%%
\resizebox{0.9\textwidth}{!}{
\begin{tabular}{ L{2.2cm} | M{1.05cm} M{1.05cm} M{1.05cm} | M{1.05cm} M{1.05cm} M{1.05cm} | M{1.05cm} M{1.05cm} M{1.05cm} M{1.05cm}}

\rowcolor{lightcyan}
\multicolumn{11}{c}{\textbf{AGIQA-3K (Alignment)}} \\
\hline

\multirow{2}{*}{\textbf{Model}}  &
\multicolumn{3}{c|}{GMC$_s$} & \multicolumn{3}{c|}{GMC$_d$} &
\multirow{2}{*}{PLCC} & \multirow{2}{*}{SRCC} & \multirow{2}{*}{KRCC}  & \multirow{2}{*}{PWRC}\\
\cline{2-4}\cline{5-7}

 & LQ & MQ & HQ & LD & MD & HD & & & & \\
\hline

\textit{CLIP-IQA}  \cite{wang2023exploring}
& \textbf{\br{0.4924}} & \textbf{\br{0.3930}} & 0.3488
& \textbf{\br{0.3280}} & \textbf{\br{0.4507}} & \textbf{\br{0.5503}}
& \textbf{\br{0.5438}} & \textbf{\br{0.4874}} & \textbf{\br{0.3365}} & \textbf{\br{12.3144}} \\

\textit{TIER}  \cite{yuan2024tier}
& 0.4129 & 0.3793 & \textit{\gr{0.3511}}
& 0.2873 & 0.4021 & 0.5028
& 0.4796 & 0.4524 & 0.3077 & 11.0080 \\

\textit{DepictQA}  \cite{you2024descriptive}
& \textit{\gr{0.4160}} & \textit{\gr{0.3824}} & \textbf{\br{0.3665}}
& \textit{\gr{0.2911}} & \textit{\gr{0.4090}} & \textit{\gr{0.5119}}
& \textit{\gr{0.4975}} & \textit{\gr{0.4525}} & \textit{\gr{0.3171}} & \textit{\gr{11.0915}} \\

\hline
\end{tabular}
}
\vspace{4pt}

%%%%%%%%%%%%%%%%%%%%%%%%%%%% AIGCIQA2023 %%%%%%%%%%%%%%%%%%%%%%%%%%%%
\resizebox{0.9\textwidth}{!}{
\begin{tabular}{ L{2.2cm} | M{1.05cm} M{1.05cm} M{1.05cm} | M{1.05cm} M{1.05cm} M{1.05cm} | M{1.05cm} M{1.05cm} M{1.05cm} M{1.05cm} }

\rowcolor{lightcyan}
\multicolumn{11}{c}{\textbf{AIGCIQA2023 (Quality)}} \\
\hline

\multirow{2}{*}{\textbf{Model}}  &
\multicolumn{3}{c|}{GMC$_s$} & \multicolumn{3}{c|}{GMC$_d$} &
\multirow{2}{*}{PLCC} & \multirow{2}{*}{SRCC} & \multirow{2}{*}{KRCC} & \multirow{2}{*}{PWRC}\\
\cline{2-4}\cline{5-7}

 & LQ & MQ & HQ & LD & MD & HD & & & & \\
\hline

\textit{CLIP-IQA}  \cite{wang2023exploring}
& 0.4659 & \textbf{\br{0.6412}} & 0.3713
& 0.2924 & 0.4727 & 0.6640
& \textbf{\br{0.7177}} & \textbf{\br{0.6887}} & \textbf{\br{0.4786}} & \textbf{\br{12.3565}} \\

\textit{TIER}  \cite{yuan2024tier}
& \textit{\gr{0.5452}} & \textit{\gr{0.6114}} & \textit{\gr{0.3908}}
& \textit{\gr{0.3189}} & \textit{\gr{0.5097}} & \textbf{\br{0.6889}}
& \textit{\gr{0.6794}} & \textit{\gr{0.6674}} & \textit{\gr{0.4628}} & \textit{\gr{11.8953}} \\

\textit{DepictQA} \cite{you2024descriptive} 
& \textbf{\br{0.5705}} & 0.5551 & \textbf{\br{0.4174}}
& \textbf{\br{0.3307}} & \textbf{\br{0.5187}} & \textit{\gr{0.6827}}
& 0.6231 & 0.6154 & 0.4371 & 11.0295 \\

\hline
\end{tabular}
}
\vspace{4pt}

%%%%%%%%%%%%%%%%%%%%%%%%%%%% AIGCIQA2023 %%%%%%%%%%%%%%%%%%%%%%%%%%%%
\resizebox{0.9\textwidth}{!}{
\begin{tabular}{ L{2.2cm} | M{1.05cm} M{1.05cm} M{1.05cm} | M{1.05cm} M{1.05cm} M{1.05cm} | M{1.05cm} M{1.05cm} M{1.05cm} M{1.05cm}}

\rowcolor{lightcyan}
\multicolumn{11}{c}{\textbf{AIGCIQA2023 (Alignment)}} \\
\hline
\multirow{2}{*}{\textbf{Model}}  &
\multicolumn{3}{c|}{GMC$_s$} & \multicolumn{3}{c|}{GMC$_d$} & \multirow{2}{*}{PLCC} & \multirow{2}{*}{SRCC} & \multirow{2}{*}{KRCC} & \multirow{2}{*}{PWRC}\\
\cline{2-4}\cline{5-7}
 & LQ & MQ & HQ & LD & MD & HD & & & & \\
\hline

\textit{CLIP-IQA} \cite{wang2023exploring}
& \textbf{\br{0.4368}} & \textit{\gr{0.4573}} & \textit{\gr{0.3100}}
& \textbf{\br{0.2538}} & \textbf{\br{0.4055}} & \textbf{\br{0.5297}}
& \textbf{\br{0.4908}} & \textbf{\br{0.5262}} & \textbf{\br{0.3673}} & \textbf{\br{9.3519}} \\

\textit{TIER}  \cite{yuan2024tier}   
& 0.3760 & \textbf{\br{0.4776}} & \textbf{\br{0.3152}}
& \textit{\gr{0.2408}} & \textit{\gr{0.3826}} & \textit{\gr{0.5118}}
& \textit{\gr{0.4659}} & \textit{\gr{0.4655}} & \textit{\gr{0.3171}} & \textit{\gr{7.0663}} \\

\textit{DepictQA} \cite{you2024descriptive}    
& \textit{\gr{0.3824}} & 0.3105 & 0.2311
& 0.1960 & 0.3200 & 0.4197
& 0.3176 & 0.3159 & 0.2194 & 5.2971 \\

\bottomrule
\end{tabular}}
\end{table*}

\begin{figure*}[!t] 
\centering 
\includegraphics[width=\textwidth]{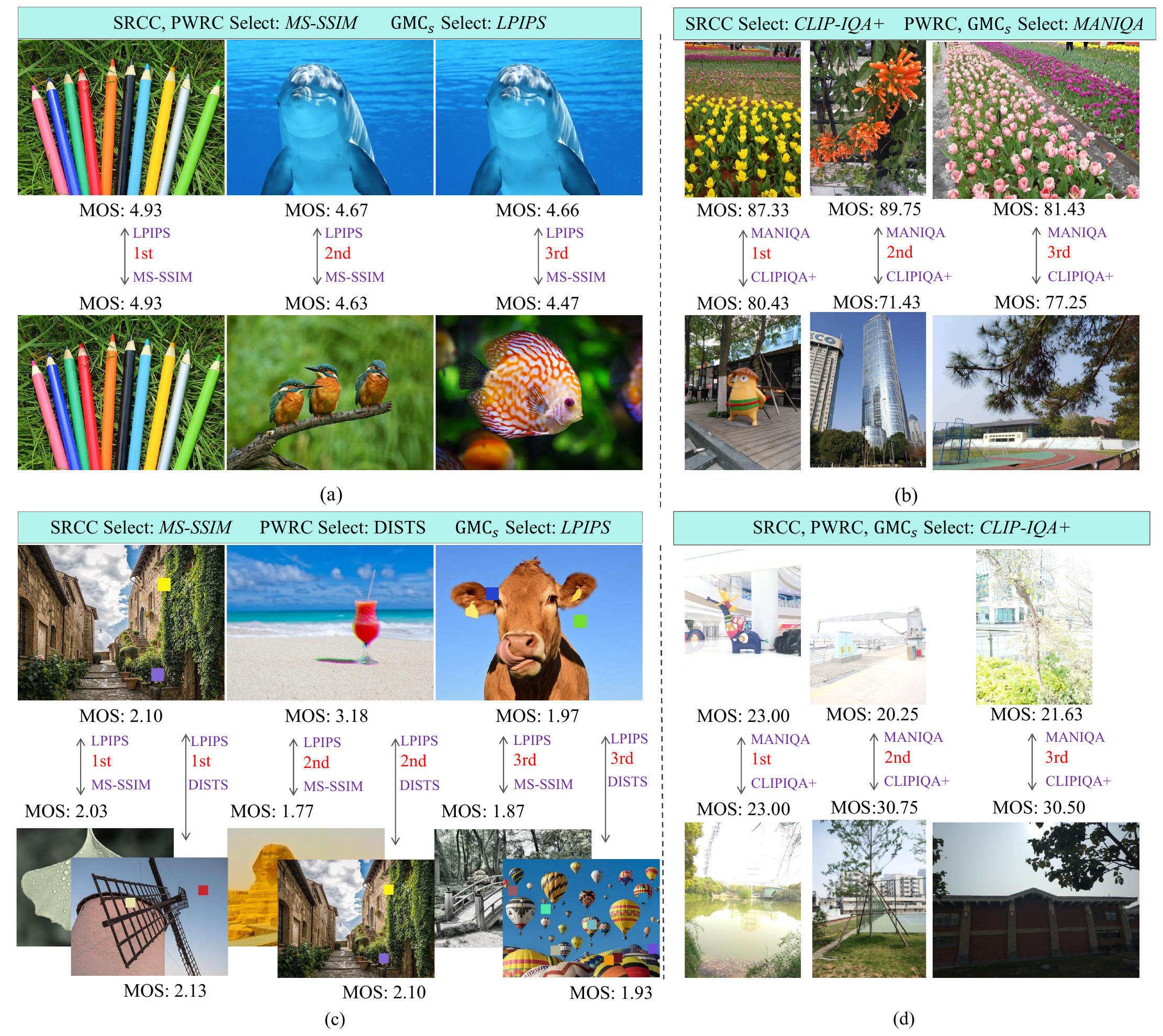} 
\caption{\rc{Comparison of high-quality images retrieved by the best FR/NR IQA models selected via GMC and the SRCC criterion.  
(a) and (b) show the top-3 images from the high-quality subsets of KADID-10K and SPAQ, respectively;  
(c) and (d) show the top-3 images from the low-quality subsets of the same datasets.  
The selected model for each subfigure is indicated in its title.}}

\label{fig:hqi} 
\end{figure*}

\begin{figure*}[t] 
\centering 
\includegraphics[width=\textwidth]{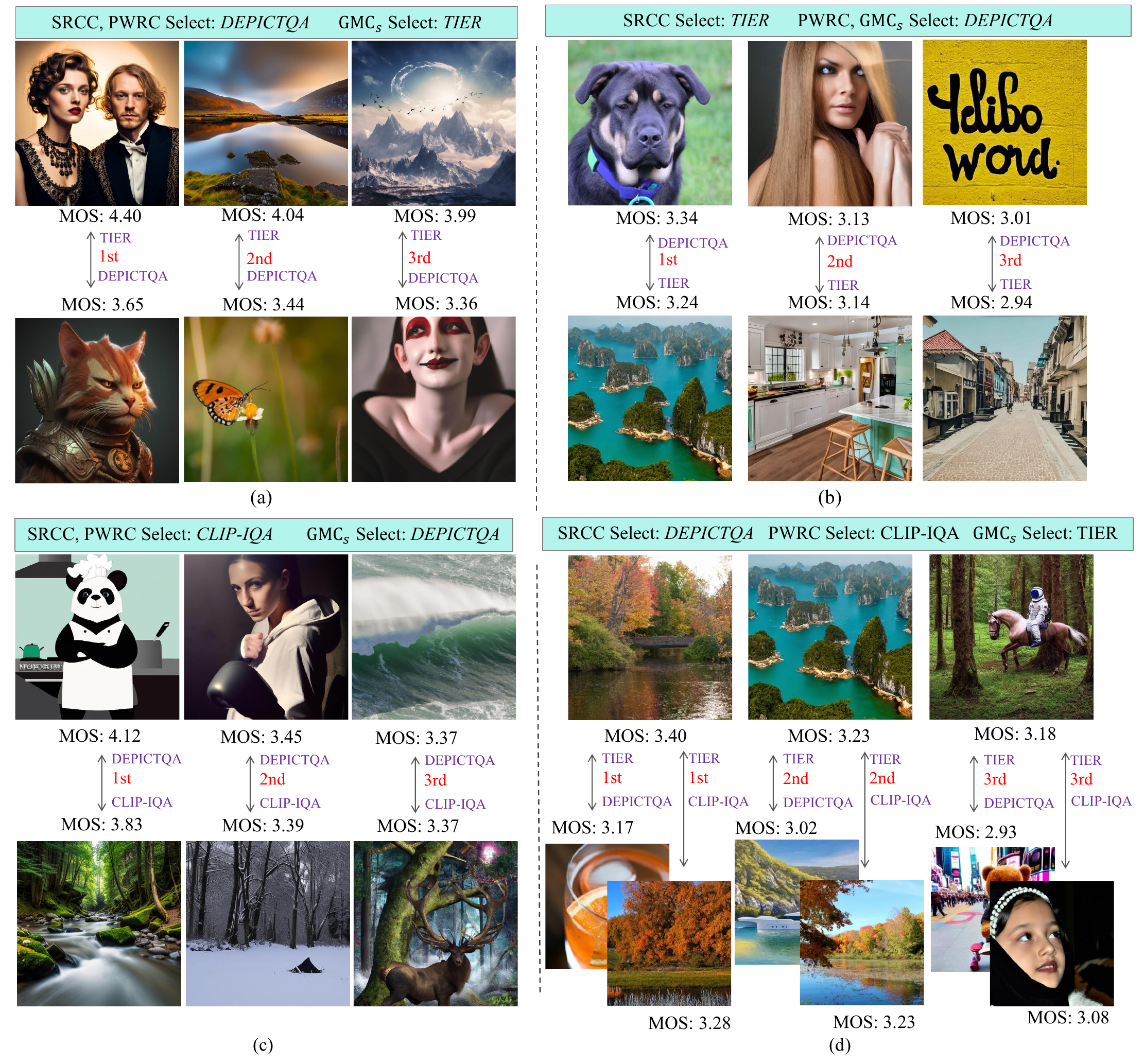} 
\caption{Comparison of images retrieved by the best IQA models selected via SRCC, PWRC, and GMC$_s$ across quality and alignment criteria.
(a) and (b) show the top-3 images from the high-quality subsets of AGIQA-3K and AIGCIQA2023, respectively;  
(c) and (d) show the top-3 images from the high-alignment subsets of the same datasets.  
The selected model for each subfigure is indicated in its title.}

\label{fig:hqi_aigc} 
\end{figure*}

\begin{figure*}[!t] 
\centering % 图形居中
\includegraphics[width=0.97\textwidth]{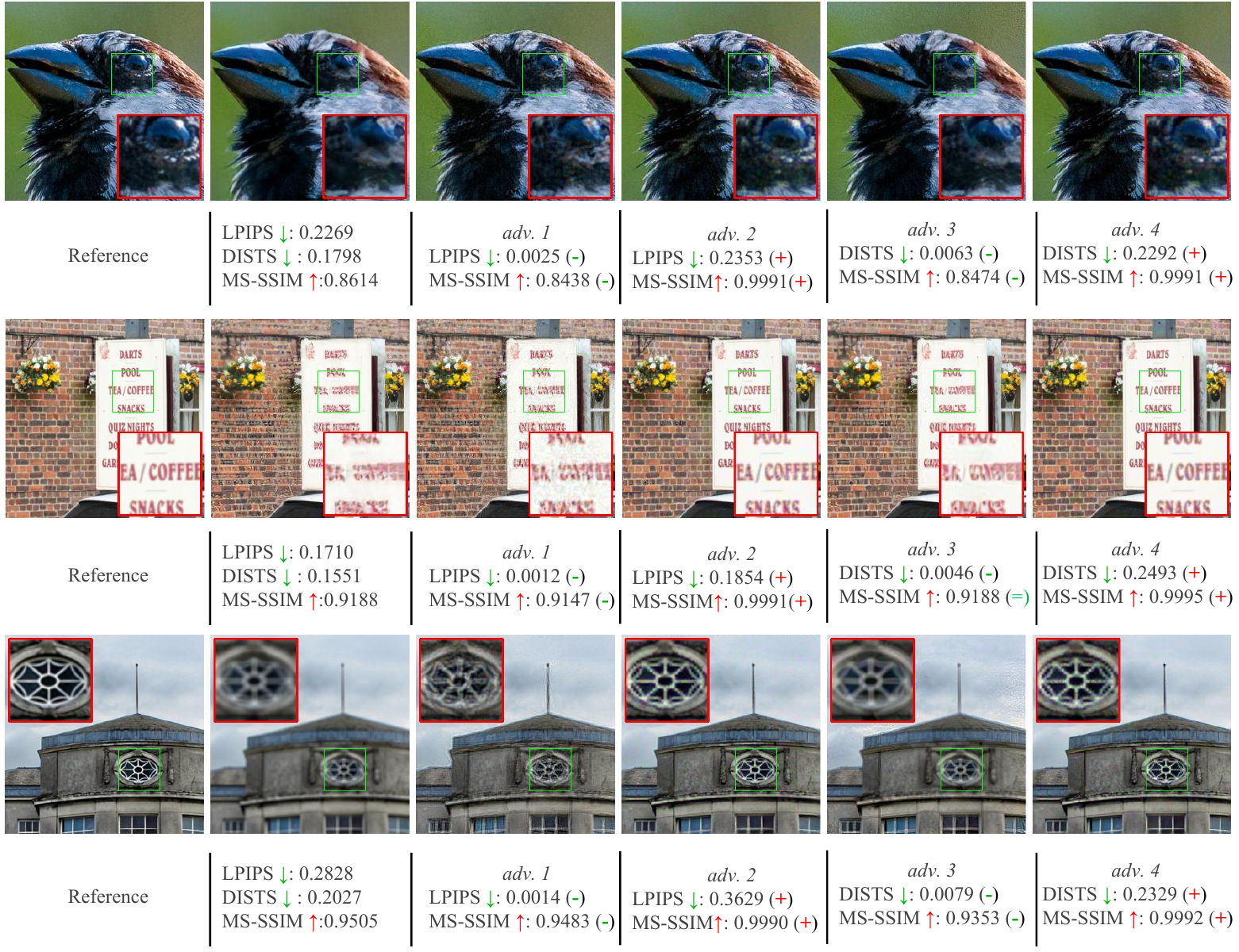} 
\vspace{-0.5em}
\caption{Qualitative comparison of adversarial image quality optimization on the PIAPL Dataset.
Columns 1 and 2 show the reference and initial distorted images.
Columns 3 and 5 present optimization results with LPIPS and DISTS as target metrics, respectively, under MS-SSIM constraints.
Columns 4 and 6 show the reverse setting, where MS-SSIM is optimized under LPIPS or DISTS constraints.} 
\label{fig:adv} 
\vspace{-1em}
\end{figure*}

\begin{figure}[htbp] 
\centering 
\includegraphics[width=0.47\textwidth]{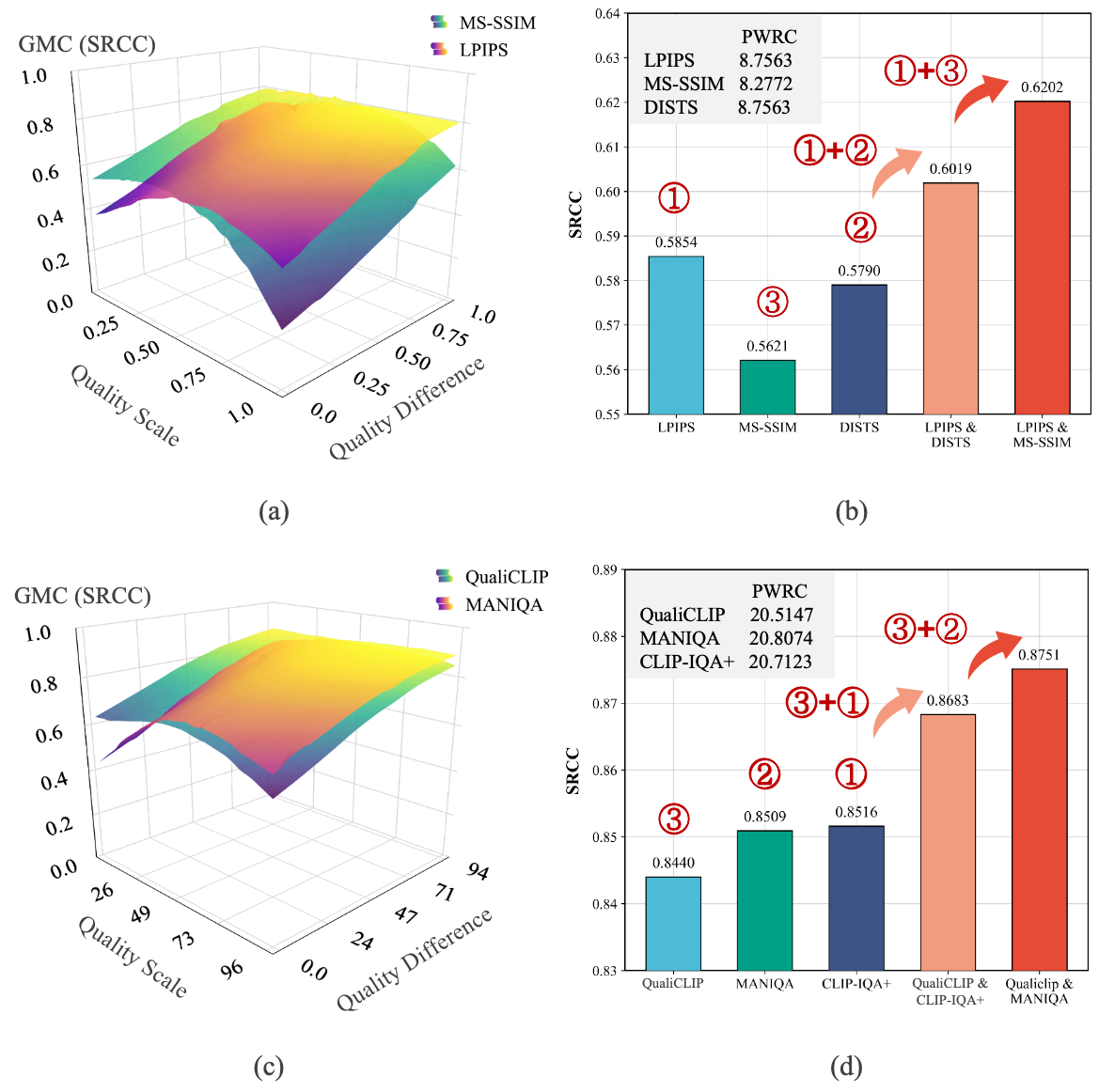} 
\vspace{-0.5em}
\caption{\rc{Comparison of IQA model integration results. (a) and (c) show the IQA model complementarity identified by our GMC on the PIPAL and SPAQ datasets, respectively. 
(b) and (d) report the SRCC  and PWRC of individual IQA models and their corresponding integrated counterparts.}}
\label{fig:cmp} 
\vspace{-2em}
\end{figure}

\begin{figure}[htp] 
\centering 
\includegraphics[width=0.47\textwidth]{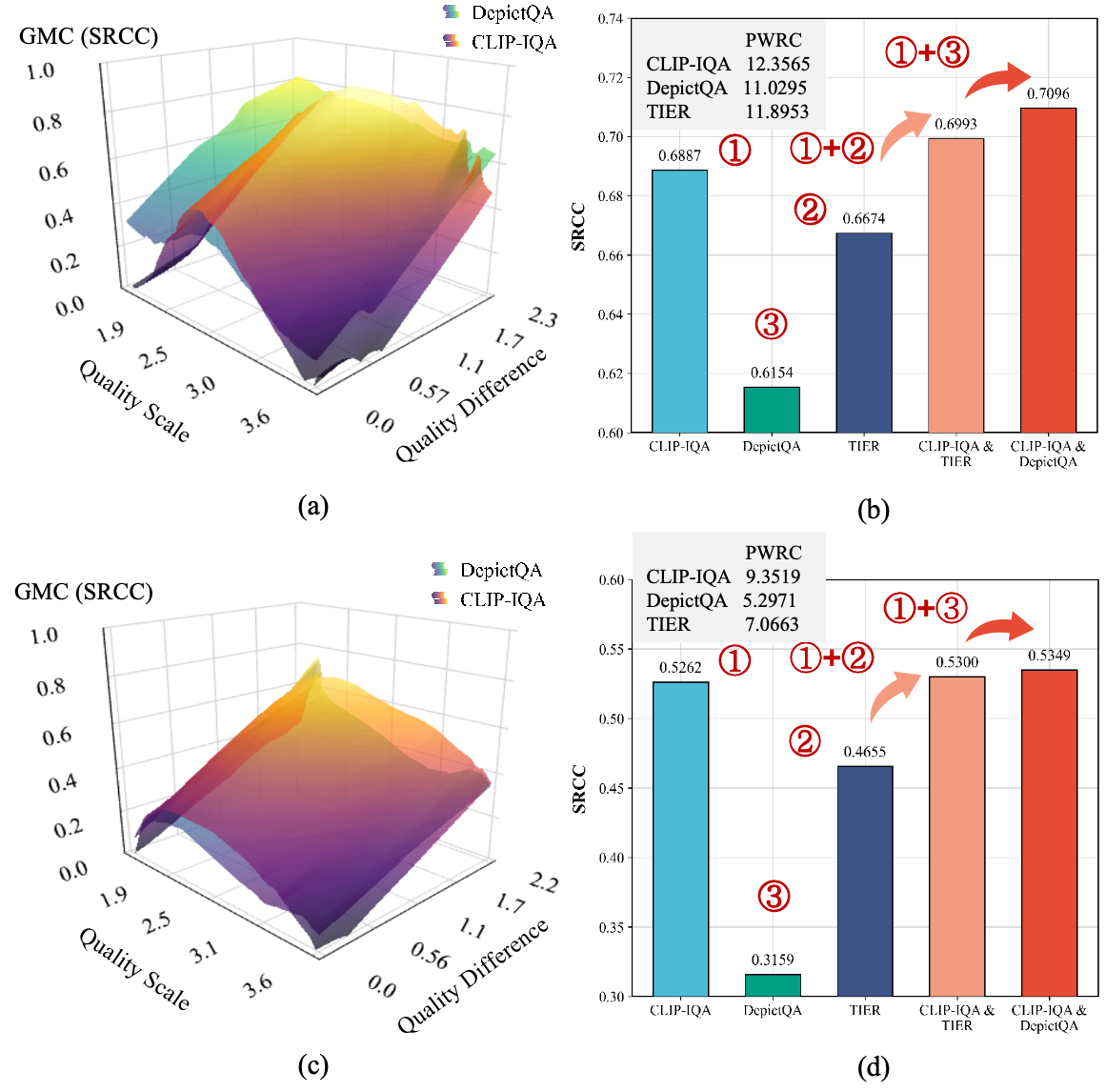} 
\caption{\rc{Comparison of AGIQA model integration results. (a) and (c) show the AGIQA model complementarity identified by our GMC on the AIGCIQA2023 datasets at two dimensions: quality and alignment, respectively. 
(b) and (d) report the SRCC and PWRC of individual IQA models and their corresponding integrated counterparts.}}
\label{fig:cmp-aigc} 
\end{figure}

\begin{figure*}[htbp] 
\centering 
\includegraphics[width=\textwidth]{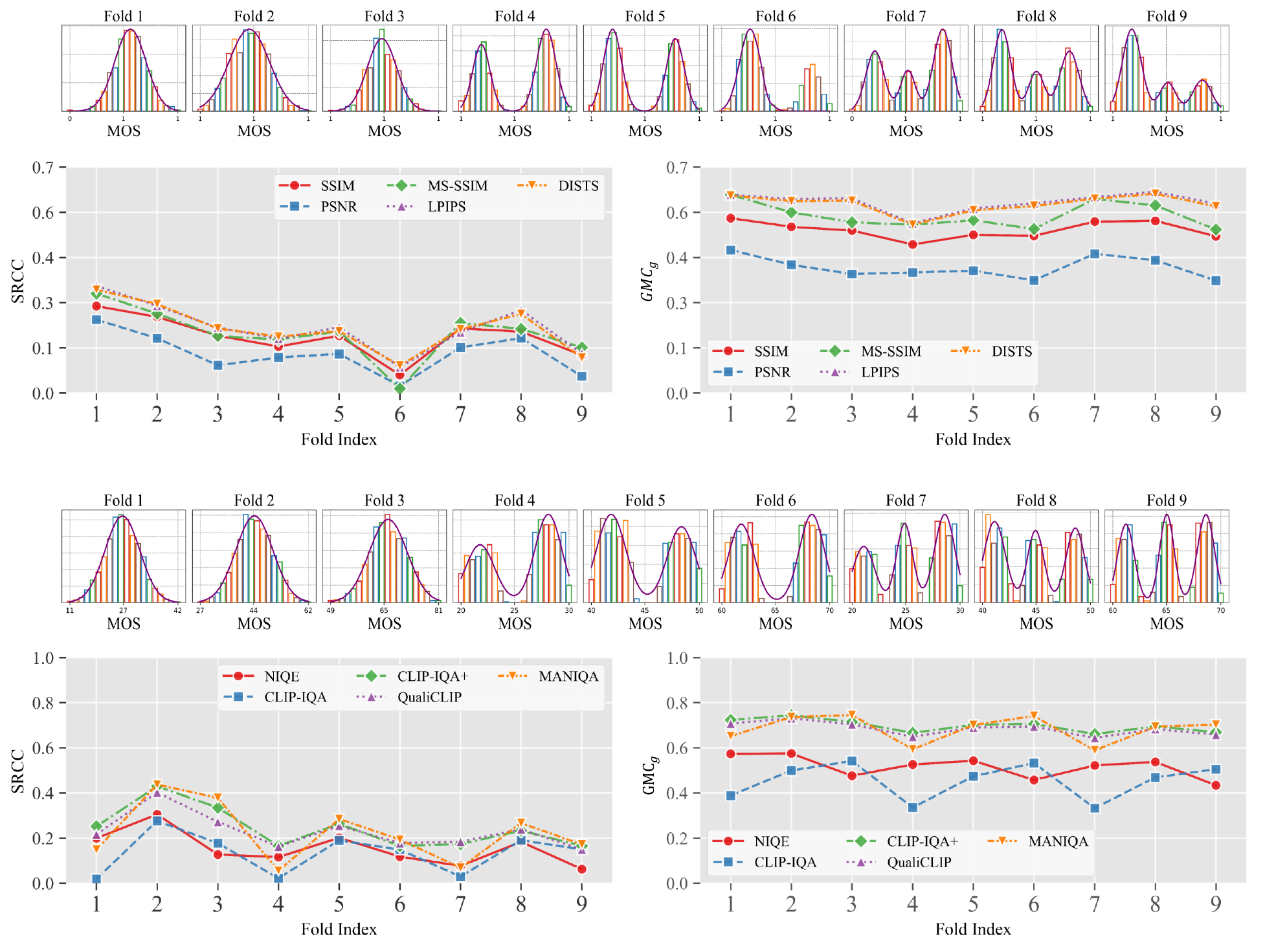} 
\caption{\rc{Robustness comparison between SRCC and $\text{GMC}_g$ across different quality distributions. 
Rows 1 and 3 present the nine sampling sets from PIPAL and SPAQ, respectively, while rows 2 and 4 show the corresponding variations in SRCC and $\text{GMC}_g$ values.}}
\label{fig:samp} 
\end{figure*}

\subsection{Applications of GMC Measure}

\noindent \textbf{(1) Insight into FR-IQA and NR-IQA Models.} 
To reveal the local prediction behaviors of existing IQA models, we visualize the joint prediction distribution of each model by our proposed GMC measure, as illustrated in Fig.~\ref{fig:3dvis}. In addition to qualitative visualization, we enable quantitative comparisons by dividing the MOS range of each dataset into three equal-width quality scales: Low Quality (LQ), Medium Quality (MQ), and High Quality (HQ). Correspondingly, we define three diagnostic intervals: Low Difference (LD), Medium Difference (MD), and High Difference (HD). For each quality scale and diagnostic interval, we compute two fine-grained diagnostic metrics, GMC$_s$ and GMC$_d$, respectively, formulated as:
\begin{equation}
    \text{GMC}_s = \frac{1}{A'} \int_{  Q^d_{\min}}^{  Q^d_{\max}} \int_{ Q^s_{1}}^{ Q^s_{2}} \hat{\Gamma}(x, y)\, dx \, dy,
\end{equation}

\begin{equation}
    \text{GMC}_d = \frac{1}{A''} \int_{  Q^s_{\min}}^{  Q^s_{\max}} \int_{ Q^d_{1}}^{ Q^d_{2}} \hat{\Gamma}(x, y)\, dx \, dy,
\end{equation}
where $[Q^s_{1}, Q^s_{2}]$ and $[Q^d_{1}, Q^d_{2}]$ define the boundaries of the selected quality scale and interval, respectively, and $A'$, $A''$ are the integration area for normalization. The comparison results of different IQA models are shown in Tab.~\ref{tab:gmc}. From the figure and table, several key insights emerge:

\begin{itemize}
\item \textit{Different IQA models exhibit distinct behaviors across MOS and $|\Delta\text{MOS}|$.} For instance, on the SPAQ dataset, MANIQA achieves consistently high GMC$_s$ scores in the medium- and high-quality regions, indicating strong robustness under perceptually favorable conditions. In contrast, CLIP-IQA+ exhibits relatively stronger performance in the low-quality and low-difference regimes, demonstrating superior discrimination capability in fine-grained quality assessment scenarios. \rc{Similar observations can also be found in AIGC scenarios (see Table~\ref{tab:gmc-aigc} and  Fig.~\ref{fig:3dvis-aigc}). On the AGIQA-3K dataset, CLIP-IQA shows superior performance in the low-quality and low-difference regions, while TIER gradually outperforms it as the quality level or quality difference increases. More importantly, we observe that PWRC tends to select the same best-performing model as GMC$_s$ in the high-quality regime in most cases. However, it fails to reveal the optimal models in other quality intervals. For instance, on the PIPAL dataset, global metrics consistently identify LPIPS as the best-performing model, whereas GMC$_s$ reveals that its performance in low-quality and low-difference regions is inferior to MS-SSIM.}\\

\item \textit{Global metrics fail to capture local behaviors.} For example, although MS-SSIM achieves a high overall SRCC on the KADID-10k dataset, its performance (GMC$_s$) in the low-quality region is notably inferior to that of DISTS. 
Similarly, while LPIPS exhibits strong performance (GMC$_d$) in the low-difference regime, it attains a lower overall PLCC than DISTS. 
These discrepancies indicate that relying solely on global correlation metrics may lead to misleading conclusions in practical evaluation scenarios. \rc{The global metrics, including PLCC, SRCC, KRCC, and PWRC, consistently designate a single model as the best-performing one across each entire dataset (\eg, TIER on AGIQA-3K Quality and CLIP-IQA on the remaining three settings). This monolithic ranking collapses heterogeneous local behaviors into a single scalar, thereby obscuring regime-specific strengths essential for scenario-aware deployment.}\\

\item \textit{Performance degradation under fine-grained quality differences.} 
Across all four datasets, a consistent performance drop can be observed when moving from high- to low-difference regimes, indicating that existing IQA models struggle to reliably discriminate subtle quality variations. For instance, on KADID-10k, the average GMC$_d$ of representative methods decreases by more than 15\% from the high-quality to low-quality region, with similar degradation trends observed on PIPAL, LIVEC, and SPAQ. 
Although some models (\eg, LPIPS or CLIP-based approaches) maintain competitive global correlation scores, their performance under low-difference conditions remains substantially weaker. 
These results suggest that strong global correlation does not necessarily imply robust fine-grained perceptual discrimination, and such limitations may be obscured when evaluation relies solely on global metrics.  \rc{Furthermore, across all alignment experiments, we observe a consistent performance degradation as the quality difference decreases (LD $<$ MD $<$ HD), indicating limited sensitivity in distinguishing subtle consistency variations. By decoupling performance along both the absolute quality and pairwise difference dimensions, GMC exposes these fine-grained limitations that remain invisible to conventional global indicators.}\\

\end{itemize}

\noindent \textbf{(2) IQA Model Selection for High-Quality Image Retrieval.} 
IQA models are commonly used to identify high-quality images across different quality regimes, such as selecting top-quality samples from generative outputs or screening out low-quality images in degraded datasets.
Standard global correlation metrics may fail to guide optimal model selection in such scenario-specific searches, particularly when the overall ranking performance of competing models is similar. To evaluate the effectiveness of our GMC measure, we construct two types of \textbf{query sets} separately for each dataset (KADID-10K and SPAQ) based on their respective MOS distributions: a high-quality set formed by images in the top quartile of MOS, and a low-quality set formed by images in the bottom quartile. \rc{For AIGC datasets (AGIQA-3K and AIGCIQA2023), we consider both the \textit{Quality} and \textit{Alignment} dimensions, and construct corresponding high-quality and high-alignment query sets using the top quartile samples in each dimension.}
\rc{For each query set, we compare the top-3 images selected by the model chosen via GMC$_s$, global SRCC, and PWRC, respectively.}

The mean MOS of the selected images is reported in Table~\ref{tab:hqi}, and the corresponding selections are visualized in Fig.~\ref{fig:hqi}. \rc{Results demonstrate that models selected by GMC$_s$ consistently identify higher-quality images in high-quality query sets and more accurately discriminate lower-quality images in low-quality query sets, outperforming those selected based on the best SRCC and PWRC in both scenarios. For AIGC datasets, the results are summarized in Table~\ref{tab:hqi_aigc} and Fig.~\ref{fig:hqi_aigc}. Results demonstrate that, across both traditional and AIGC datasets, models selected by GMC$_s$ consistently outperform those chosen based on SRCC or PWRC. Specifically, GMC$_s$ enables more accurate identification of high-quality images and more reliable discrimination of lower-quality samples in traditional IQA datasets, while in AIGC scenarios, it further facilitates stable retrieval of images with higher perceptual quality or better semantic alignment.}\\

\begin{table}[htbp]
\centering
\caption{
\rc{Performance comparison for high-quality image selection on traditional IQA datasets when the best model is selected by SRCC, PWRC, or $\text{GMC}_s$. Blue star ($\color{blue}{\bigstar}$) denotes the model selected by SRCC, gray star ($\color{gray}{\bigstar}$) denotes the model selected by PWRC, and red star ($\color{red}{\bigstar}$) denotes the model selected by $\text{GMC}_s$.}
}

\begin{tabular}{l l l c c}
\toprule
Query Set & Dataset & Model & Selection & Mean MOS \\
\midrule
\multirow{4}{*}{High-quality} 
 & \multirow{2}{*}{KADID-10k} & MS-SSIM & $\color{blue}{\bigstar}$ $\color{gray}{\bigstar}$ & 4.68 \\
 & & \textbf{LPIPS} & $\color{red}{\bigstar}$ & \textbf{4.75} \\
 \cmidrule(l){2-5}
 & \multirow{2}{*}{SPAQ} & CLIP-IQA+ & $\color{blue}{\bigstar}$ & 76.37 \\
 & & \textbf{MANIQA} & $\color{gray}{\bigstar}$ $\color{red}{\bigstar}$ & \textbf{86.17} \\
\midrule
\multirow{5}{*}{Low-quality} 
 & \multirow{3}{*}{KADID-10k} & MS-SSIM & $\color{blue}{\bigstar}$ & 1.89 \\
 & & DISTS & $\color{gray}{\bigstar}$ & 2.05 \\
 & & \textbf{LPIPS} & $\color{red}{\bigstar}$ & \textbf{2.42} \\
  \cmidrule(l){2-5}
 & \multirow{2}{*}{SPAQ} & MANIQA & - & 21.63 \\
 & & \textbf{CLIP-IQA+} & $\color{blue}{\bigstar}$ $\color{gray}{\bigstar}$ $\color{red}{\bigstar}$ & \textbf{28.08} \\
\bottomrule
\end{tabular}
\label{tab:hqi}
\end{table}

% \begin{table}[t]
% \centering
% \caption{Performance comparison for high-quality image selection when the best IQA model is selected by  SRCC or $\text{GMC}_s$. 
% Red star ($\color{red}{\bigstar}$) denotes the model selected by SRCC, and blue star ($\color{blue}{\bigstar}$) denotes the model selected by $\text{GMC}_s$.}

% \begin{tabular}{l l l c c}
% \toprule
% Query Set & Dataset & Model & Selection & Mean MOS \\
% \midrule
% \multirow{4}{*}{High-quality} 
%  & \multirow{2}{*}{KADID-10k} & MS-SSIM & $\color{blue}{\bigstar}$ & 4.68 \\
%  & & \textbf{LPIPS} & $\color{red}{\bigstar}$ & \textbf{4.75} \\
%  \cmidrule(l){2-5}
%  & \multirow{2}{*}{SPAQ} & CLIP-IQA+ & $\color{blue}{\bigstar}$ & 76.37 \\
%  & & \textbf{MANIQA} & $\color{red}{\bigstar}$ & \textbf{86.17} \\
% \midrule
% \multirow{4}{*}{Low-quality} 
%  & \multirow{2}{*}{KADID-10k} & MS-SSIM & $\color{blue}{\bigstar}$ & 1.89 \\
%  & & \textbf{LPIPS} & $\color{red}{\bigstar}$ & \textbf{2.42} \\
%   \cmidrule(l){2-5}
%  & \multirow{2}{*}{SPAQ} & MANIQA & - & 21.63 \\
%  & & \textbf{CLIP-IQA+} & $\color{blue}{\bigstar}$ $\color{red}{\bigstar}$ & \textbf{28.08} \\
% \bottomrule
% \end{tabular}
% \label{tab:hqi}
% \end{table}

\begin{table}[t]
\centering
\caption{
\rc{Performance comparison for high-quality and high-alignment image selection on AIGC datasets when the best model is selected by SRCC, PWRC, or $\text{GMC}_s$. Blue star ($\color{blue}{\bigstar}$) denotes the model selected by SRCC, gray star ($\color{gray}{\bigstar}$) denotes the model selected by PWRC, and red star ($\color{red}{\bigstar}$) denotes the model selected by $\text{GMC}_s$.
}}
\begin{tabular}{l l l c c}
\toprule
Query Set & Dataset & Model & Selection & Mean MOS \\
\midrule
\multirow{4}{*}{Quality} 
 & \multirow{2}{*}{AGIQA-3K} & DepictQA & $\color{blue}{\bigstar}$ $\color{gray}{\bigstar}$ & 3.48 \\
 & & \textbf{TIER} & $\color{red}{\bigstar}$ & \textbf{4.14} \\
 \cmidrule(l){2-5}
 & \multirow{2}{*}{AIGCIQA2023} & TIER & $\color{blue}{\bigstar}$ & 3.11 \\
 & & \textbf{DepictQA} & $\color{gray}{\bigstar}$ $\color{red}{\bigstar}$ & \textbf{3.16} \\
\midrule
\multirow{5}{*}{Alignment} 
 & \multirow{3}{*}{AGIQA-3K} & CLIP-IQA & $\color{blue}{\bigstar}$ $\color{gray}{\bigstar}$ & 3.53 \\
 & & \textbf{DepictQA} & $\color{red}{\bigstar}$ & \textbf{3.65} \\
  \cmidrule(l){2-5}
 & \multirow{3}{*}{AIGCIQA2023} & DepictQA & $\color{blue}{\bigstar}$ & 3.04 \\
 & & CLIP-IQA & $\color{gray}{\bigstar}$ & 3.20 \\
 & & \textbf{TIER} & $\color{red}{\bigstar}$ & \textbf{3.27} \\
\bottomrule
\end{tabular}
\label{tab:hqi_aigc}
\end{table}

\noindent \textbf{(3) Fine-Grained Quality Optimization.}
To assess the practical impact of GMC in optimization settings, we consider fine-grained quality optimization, where an IQA model is used as an objective function to guide the iterative refinement of enhanced (distorted) images. In this regime, perceptual differences between reference and optimized images are subtle, and effective optimization depends on the ability of the IQA model to discriminate small yet meaningful quality variations.

Table~\ref{tab:gmc} indicates that MS-SSIM consistently outperforms LPIPS and DISTS in terms of discriminative capability in the low quality-difference (LD) regime.
To validate this finding, we design an \emph{adversarial optimization} based comparison framework on the PIPAL dataset.

Specifically, given a reference image \(y\) and an initial distorted image \(x_0\)
that exhibits a slight quality degradation relative to \(y\), a target IQA model is optimized to enhance image quality (minimizing its loss), whereas a competing metric is adversarially constrained to prevent improvement.
Accordingly, we consider two complementary optimization settings,
which are formulated as the following constrained problems:

\begin{equation}
\min_{x}   f_{\text{MS-SSIM}}(x, y), \:
\text{s.t.}  \;
0 \;\le\; f_{\text{LPIPS}}(x, y) - f_{\text{LPIPS}}(x_0, y) \;\le\; \epsilon,
\end{equation}
and
\begin{equation}
\min_{x}   f_{\text{LPIPS}}(x, y), \: 
\text{s.t.} \;
0 \;\le\; f_{\text{MS-SSIM}}(x, y) - f_{\text{MS-SSIM}}(x_0, y) \;\le\; \epsilon .
\end{equation}
Here, \( f_{\text{MS-SSIM}} \) and \( f_{\text{LPIPS}} \) denote the corresponding loss functions
of perceptual quality metrics,
and \( \epsilon \) controls the tolerance threshold
that bounds the allowable degradation of the adversarial metric.
In practice, we optimize the above objectives using their unconstrained penalized forms,
where violations of the adversarial constraints are softly penalized
through hinge-based regularization.
The threshold \( \epsilon \) is introduced to prevent the optimization
from exploiting trivial shortcuts,
where the overall loss could be reduced
by excessively deteriorating the adversarial metric
without yielding a genuine improvement in the target metric.

As shown in Fig.~\ref{fig:adv}, when optimized against LPIPS or DISTS as adversarial metrics, improvements in MS-SSIM consistently translate into perceptual quality enhancement. In contrast, when MS-SSIM serves as the defender metric, optimizing LPIPS or DISTS leads to visually unsatisfactory results, even though their metric values improve.
These observations align with the GMC analysis, confirming that models exhibiting higher GMC scores under low quality-difference regimes possess superior robustness and reliability for fine-grained quality optimization tasks.\\

\noindent \textbf{(4) Model Integration for Enhanced IQA Performance.}  
Beyond quality optimization, GMC can also reveal complementary relationships among IQA models, enabling performance improvement through model integration. To demonstrate this, we conduct experiments on both FR- and NR-IQA datasets. On the FR-IQA dataset PIPAL, three representative models are selected: a traditional hand-crafted feature model (MS-SSIM) and two deep feature models (LPIPS and DISTS), with LPIPS achieving the best individual performance. \rc{Notably, although both SRCC and PWRC suggest that DISTS outperforms MS-SSIM in terms of overall performance, GMC reveals that MS-SSIM provides more complementary information to LPIPS,}
%Guided by GMC, MS-SSIM exhibits higher complementarity with LPIPS than with DISTS, 
particularly on low-quality and low-difference subsets, as indicated by the larger GMC values in Table~\ref{tab:gmc}. This suggests that integrating LPIPS with MS-SSIM is expected to yield superior performance, despite DISTS exhibiting higher SRCC performance.

The integration is formulated as a weighted combination of the metrics, taking into account their opposite polarity: lower LPIPS and DISTS values indicate better quality, whereas higher MS-SSIM values indicate better quality. Specifically, we define
\begin{equation}
\begin{aligned}
V_{Inter1} &= (1 - V_\text{LPIPS}) + V_\text{MS-SSIM},\\
V_{Inter2} &= (1 - V_\text{DISTS}) + V_\text{MS-SSIM},
\end{aligned}
\end{equation}
where $V$ denotes the quality values predicted by each model.
Experimental results in Fig.~\ref{fig:cmp} validate that the ``LPIPS \& MS-SSIM" integration achieves SRCC = 0.6202, compared with ``LPIPS \& DISTS" (SRCC = 0.6019), resulting in a relative improvement of 3.0\%.

A similar trend also holds on the NR-IQA dataset SPAQ: integrating the baseline QualiCLIP model with MANIQA, as revealed by GMC, outperforms the SRCC-selected CLIP-IQA+ model, demonstrating that the complementary integration strategy generalizes across both FR and NR scenarios. 

\rc{We further extend this analysis to the AIGCIQA2023 dataset. The results are presented in Fig.~\ref{fig:cmp-aigc}. In the quality dimension, we consider integrating CLIP-IQA with other candidate models. Although both SRCC and PWRC indicate that TIER outperforms DepictQA globally, GMC analysis reveals a different insight: CLIP-IQA achieves the best performance in the medium-quality region, while DepictQA performs better in low- and high-quality regions, indicating stronger complementarity between CLIP-IQA and DepictQA. Guided by this observation, we perform model integration and find that the combination of CLIP-IQA and DepictQA achieves a higher SRCC than that of CLIP-IQA and TIER.}

\rc{An analogous trend is observed regarding the Alignment dimension. Although TIER achieves higher global SRCC and PWRC scores than DepictQA, GMC surfaces reveal that DepictQA exhibits stronger complementary behavior with CLIP-IQA across different alignment regimes. Consequently, integrating CLIP-IQA with DepictQA yields better performance than integrating it with TIER.}

\rc{These results indicate that GMC not only facilitates informed model selection but also provides critical guidance for identifying complementary models, enabling more effective integration across traditional and AIGC IQA scenarios.}
\\

\noindent \textbf{(5) Robustness Under Varied Sampling Distributions.} 
\label{sec:42}
In real-world scenarios, image datasets often exhibit imbalanced or non-uniform quality distributions. Under such conditions, a robust evaluation metric is crucial for ensuring fair and reliable comparisons among IQA models. To demonstrate the robustness of $\text{GMC}_g$ under varying sampling distributions, we conduct systematic experiments on two widely used benchmarks, PIPAL and SPAQ, involving a diverse set of both FR and NR IQA models.

To simulate realistic distributional shifts, we adopt a Gaussian probability–weighted sampling strategy to construct imbalanced subsets from each dataset. For each dataset, three sampling variants: unimodal, bimodal, and trimodal, are generated. Each variant further produces three sub-variants, yielding a total of nine samples for analysis. As the MOS scales differ across datasets, all sampled scores are normalized to the range $[0, 100]$, partitioned into bins of size 1, and the number of samples in each bin is counted. The resulting sampling distributions are shown in Fig.~\ref{fig:samp}.

We evaluate all considered IQA models on the constructed subsets using both the global SRCC metric and the proposed $\text{GMC}_g$. As shown in Fig.~\ref{fig:samp}, $\text{GMC}_g$ consistently exhibits lower variability than SRCC across the nine subsets for every model–dataset pair, demonstrating its enhanced robustness to sampling-induced distributional bias.

\begin{figure*}[htbp]
\centering 
\includegraphics[width=\textwidth]{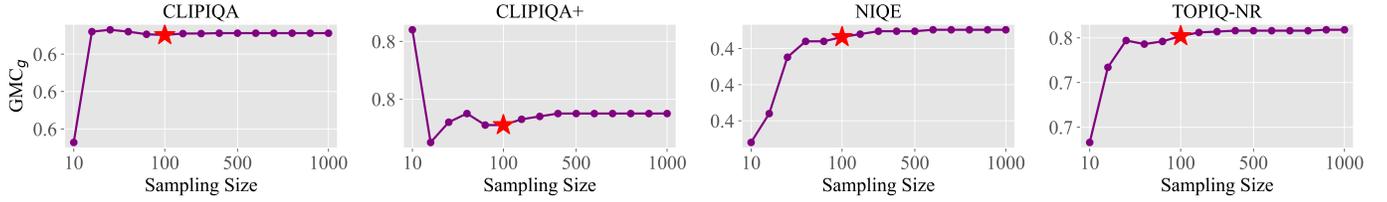} 
\caption{Variation of $\text{GMC}_g$ across IQA models versus sampling size under Latin Hypercube Sampling.} 
\label{tab:lhs-samp} 
\end{figure*}

\begin{figure*}[htbp]
\centering 
\includegraphics[width=\textwidth]{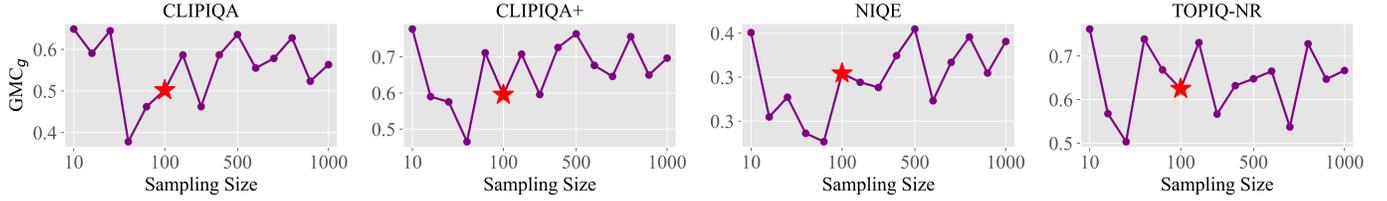} 
\caption{Variation of $\text{GMC}_g$ across IQA models versus sampling size under random sampling.} 
\label{tab:rand-samp} 
\end{figure*}

\subsection{Ablation Study}\label{sec:as}
\noindent\textbf{Study on Kernel-Smoothed Density Estimation.}
\rc{To validate the importance of kernel-smoothed density estimation, we conduct an ablation study comparing the full GMC against a variant that replaces $D(q_i)$ in Eqn.~(16) with the reciprocal of the raw sample frequency. For each sampling instance, both surfaces are re-estimated directly from the sampled data, and GMC is computed via global integration over the instance-specific fitted surface. We conduct 50 extreme sampling trials. The hybrid sampling strategy combines: (i) concentrated distributions with small variances; (ii) heuristic bias toward data-sparse regions via histogram-guided Gaussian center placement; and (iii) fully random mixtures with broad variances from a Dirichlet distribution. Results in Table~\ref{tab:kernel-smooth} show that the kernel-smoothed variant achieves lower variance, demonstrating superior robustness under distributional perturbations.} \\

%To validate the importance of the kernel-smoothed density estimation in improving the robustness of our proposed GMC metric under imbalanced MOS distributions, we conducted an ablation study by comparing the full GMC with a variant that omits the kernel-smoothing step. In particular, the density term $\mathcal{D}(q_i)$ in Eqn.~(\ref{eq:weight}) is replaced with the reciprocal of the raw sample frequency of the bin to which $q_i$ belongs.   Table~\ref{tab:kernel-smooth} reports the variances computed using both variants of GMC under the nine sampling results described in Sec.~\ref{sec:42} (5). The results clearly show that the original GMC with kernel-smoothed density consistently achieves lower variance, indicating higher robustness against distributional shifts.

\noindent\textbf{Study on Sampling Size in 3D Performance Surface Modeling.}
To evaluate the impact of sampling density on the stability of our 3D performance surface modeling, we conduct an ablation study on the number of sampling points used in the Latin Hypercube Sampling (LHS) scheme.
We vary the number of LHS sampling points ($K$ in Eqn.~(\ref{eq:lhs})) from 10 to 1,000 and compute the resulting performance GMC$_g$ for each setting. For comparison, we also report results using naïve random sampling with the same sample sizes. The results are summarized in Fig.~\ref{tab:lhs-samp} and \ref{tab:rand-samp}.

It is evident that random sampling leads to high variance and unstable evaluations, even when the sample size is sufficiently large. In contrast, LHS achieves stable and reliable estimates with significantly fewer samples. Notably, when the number of LHS samples exceeds 100, the computed metrics converge and remain stable, indicating that LHS not only enhances sampling efficiency but also ensures robustness in surface modeling. These findings validate the use of LHS as an effective and practical strategy for performance surface construction, striking a favorable balance between computational efficiency and statistical reliability.

\begin{table}[t]
    \centering
    \caption{\rc{Effect of kernel smoothing on $\text{GMC}_g$ variance across IQA models.}}

    \label{tab:kernel-smooth}
    \renewcommand{\arraystretch}{1.1}

    \begin{tabular}{
        >{\centering\arraybackslash}m{0.12\linewidth}
        >{\centering\arraybackslash}m{0.13\linewidth}
        p{0.20\linewidth}
        >{\centering\arraybackslash}m{0.12\linewidth}
        >{\centering\arraybackslash}m{0.09\linewidth}
    }
        \toprule
        & Kernel & Model & Standard Deviation  & Average \\
        \midrule

        % -------- PIPAL --------
        \multirow{10}{*}{\rotatebox{90}{PIPAL}}
         & \multirow{5}{*}{\textcolor{red}{\ding{55}}}
            & PSNR      & 0.1438 & \multirow{5}{*}{\centering 0.1664} \\
         &  & SSIM      & 0.1717 & \\
         &  & MS-SSIM   & 0.1932 & \\
         &  & LPIPS     & 0.1610 & \\
         &  & DISTS     & 0.1621 & \\
        \cmidrule(l){2-5}
         & \multirow{5}{*}{\textcolor{green}{\ding{51}}}
            & PSNR      & 0.1476 & \multirow{5}{*}{\centering \textbf{0.1651}} \\
         &  & SSIM      & 0.1678 & \\
         &  & MS-SSIM   & 0.1924 & \\
         &  & LPIPS     & 0.1601 & \\
         &  & DISTS     & 0.1574 & \\

        \midrule

        % -------- SPAQ --------
        \multirow{10}{*}{\rotatebox{90}{SPAQ}}
         & \multirow{4}{*}{\textcolor{red}{\ding{55}}}
            & NIQE   & 0.1445 & \multirow{5}{*}{\centering 0.1778} \\
         & & CLIP-IQA  & 0.1966 & \\
         & & CLIP-IQA+ & 0.1665 & \\
         & & QualiCLIP  & 0.1672 & \\
         & & MANIQA  & 0.2141 & \\
        \cmidrule(l){2-5}
         & \multirow{4}{*}{\textcolor{green}{\ding{51}}}
            & NIQE   & 0.1484 & \multirow{5}{*}{\centering \textbf{0.1737}} \\
         & & CLIP-IQA  & 0.1777 & \\
         & & CLIP-IQA+  & 0.1615 & \\
         & & QualiCLIP    & 0.1671 & \\
         & & MANIQA  & 0.2138 & \\
        
        \midrule
        
          % -------- SPAQ --------
        \multirow{6}{*}{\rotatebox{90}{AIGCIQA2023}}
         & \multirow{3}{*}{\textcolor{red}{\ding{55}}}
            & CLIP-IQA   & 0.1928 & \multirow{3}{*}{\centering 0.1783} \\
         & & TIER  & 0.1665 & \\
         & & DepictQA & 0.1758 & \\
        \cmidrule(l){2-5}
         & \multirow{3}{*}{\textcolor{green}{\ding{51}}}
            & CLIP-IQA   & 0.1912 & \multirow{3}{*}{\centering \textbf{0.1626}} \\
         & & TIER  & 0.1551 & \\
         & & DepictQA  & 0.1416 & \\

        \bottomrule
    \end{tabular}
\end{table}

\section{Conclusion}
In this work, we have revisited the fundamental problem of IQA model evaluation and argued that conventional global correlation metrics are inherently limited in their ability to reveal fine-grained model behavior and are highly sensitive to quality distribution biases. To address these challenges, we proposed a GMC framework that extends classical correlation analysis with granularity-aware modulation and distribution-aware regularization. By constructing a continuous correlation surface over quality score and quality difference, GMC enables a structured and interpretable characterization of IQA performance. 

The GMC analysis reveals that a critical {performance decoupling} where strong global correlation frequently fails to translate into fine-grained discrimination—particularly in low-difference regimes. This finding suggests that future IQA model design must move beyond broad statistical alignment and prioritize localized sensitivity to satisfy the stringent requirements of fine-grained perceptual tasks.  Furthermore, GMC advocates for a shift from searching for a "universally optimal" metric to identifying "scenario-appropriate" models. By visualizing the performance landscape, we provide a principled basis for model deployment and optimization tailored to specific perceptual constraints. Finally, by exposing how IQA models inherit and manifest biases from their training distributions, GMC could serve as a diagnostic lens to guide the acquisition of balanced datasets and the development of distribution-robust architectures. In essence, GMC transforms evaluation from a passive post-hoc measurement into an active, diagnostic bridge that links model behavior, dataset composition, and real-world deployment requirements.

\bibliographystyle{IEEEtran}\
\small{
\bibliography{ref}}

\begin{IEEEbiography}[{\includegraphics[width=1in,height=1.25in, clip,keepaspectratio]{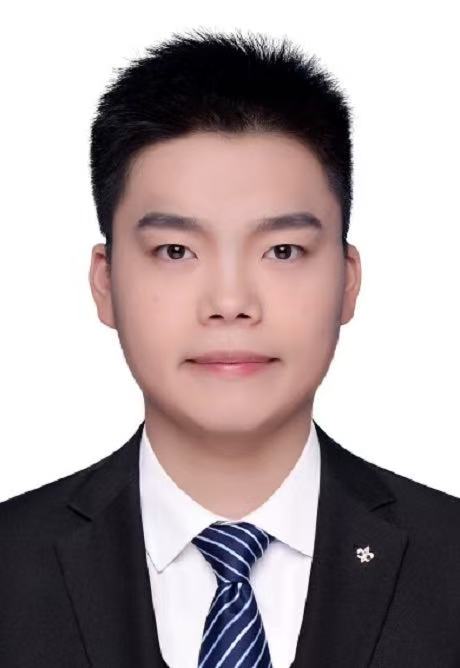}}]{Baoliang Chen} (Member, IEEE) received the Ph.D. degree in Computer Science from the City University of Hong Kong, Hong Kong, in 2022. From 2024 to 2026, he was an Associate Professor with South China Normal University. Since 2026, he has been a Postdoctoral Research Fellow with Nanyang Technological University, Singapore. His research interests include image/video quality assessment and transfer learning.
\end{IEEEbiography}

\begin{IEEEbiography}[{\includegraphics[width=1in,height=1.25in, clip,keepaspectratio]{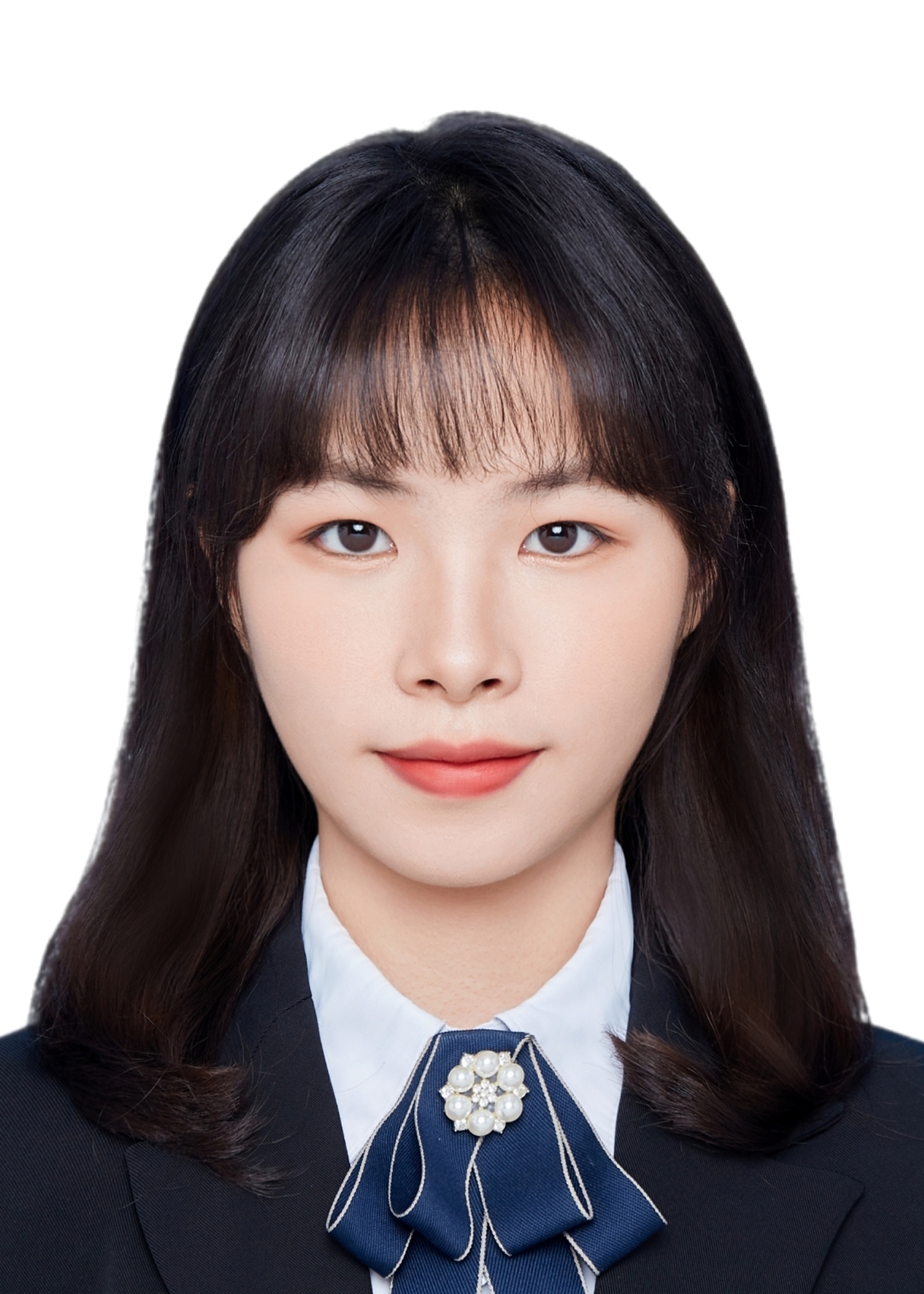}}]{Danni Huang} is currently pursuing the B.E. degree with the Department of Computer Science, South China Normal University, China. Her research interests include image quality assessment and image quality enhancement.
\end{IEEEbiography}

\begin{IEEEbiography}[{\includegraphics[width=1in,height=1.25in, clip,keepaspectratio]{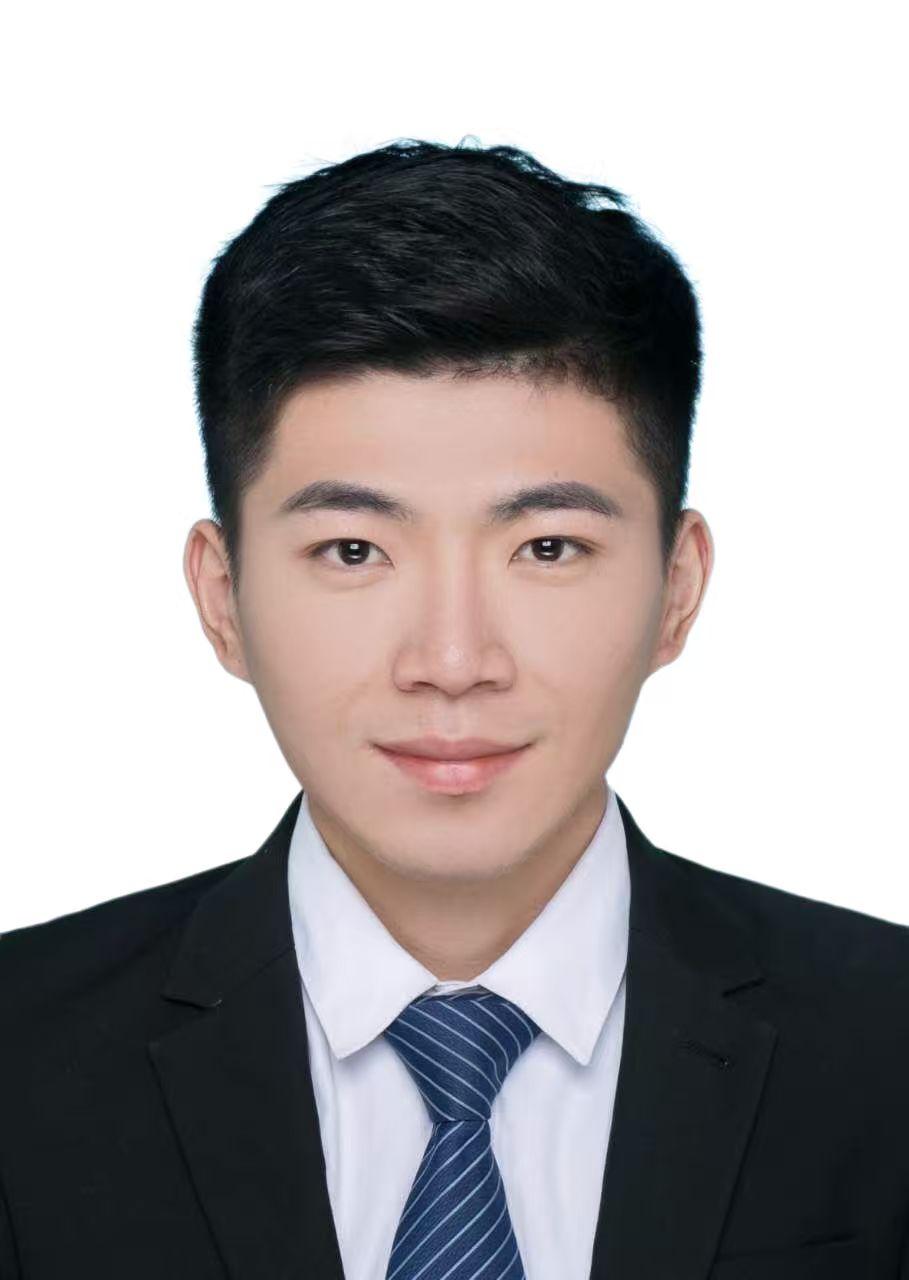}}]{Hanwei Zhu} (Member, IEEE) received  Ph.D. degree in computer science from the City University of Hong Kong, Hong Kong, in 2025.  He is currently a research scientist with the Alibaba-NTU Global e-Sustainability CorpLab (ANGEL) at Nanyang Technological University. His research interests include perceptual image processing, computational vision, and computational photography.
\end{IEEEbiography}

\begin{IEEEbiography}[{\includegraphics[width=1in,height=1.25in, clip,keepaspectratio]{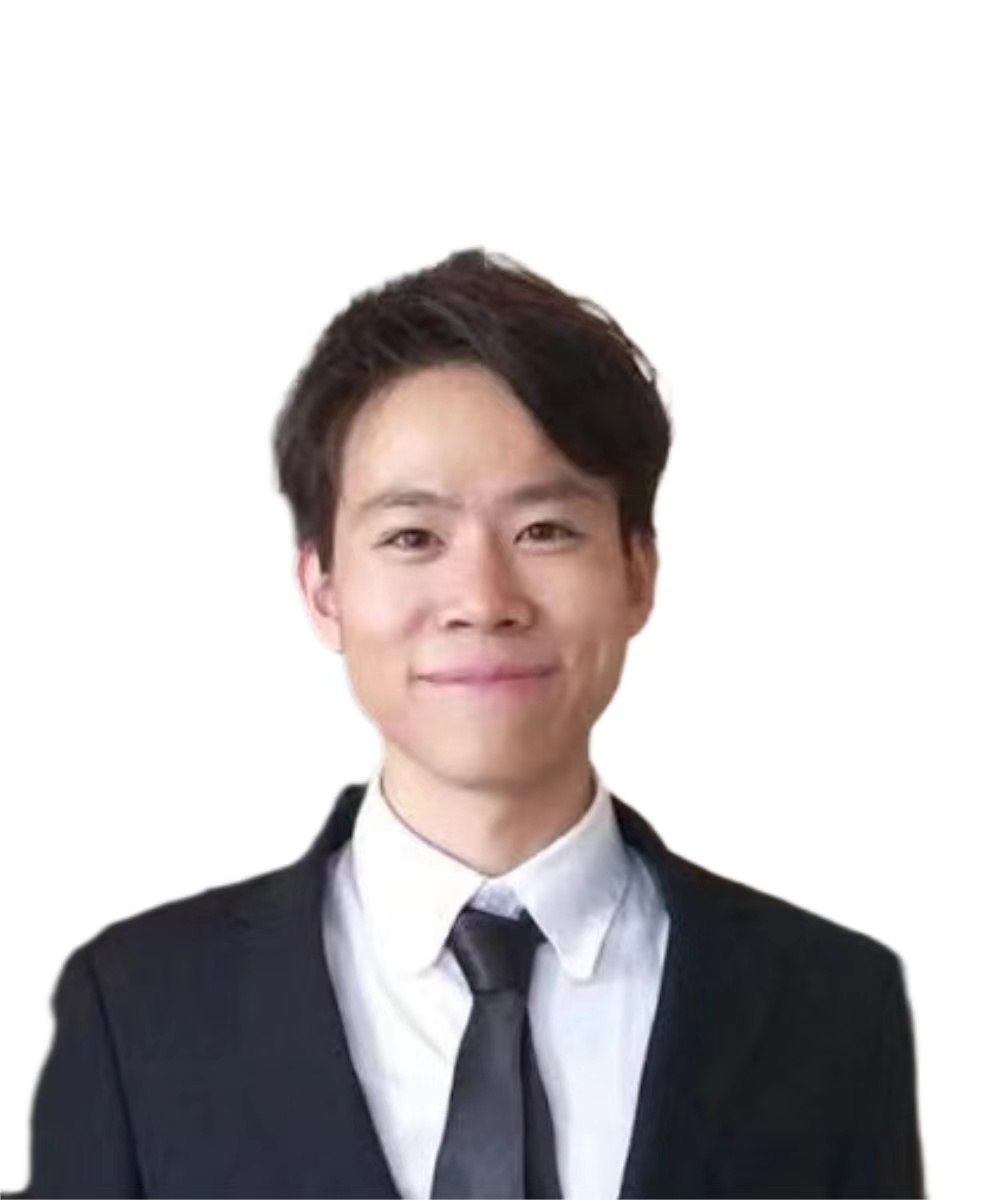}}]{Lingyu Zhu} (Member, IEEE) received his Ph.D. degree in computer science from the City University of Hong Kong, Hong Kong SAR, China, in 2024. He is currently a postdoctoral researcher with the Department of Computer Science at City University of Hong Kong. His research interests include image/video compression, image/video enhancement, image/video quality assessment, and deep learning.
\end{IEEEbiography}

\begin{IEEEbiography}[{\includegraphics[width=1in,height=1.25in,clip,keepaspectratio]{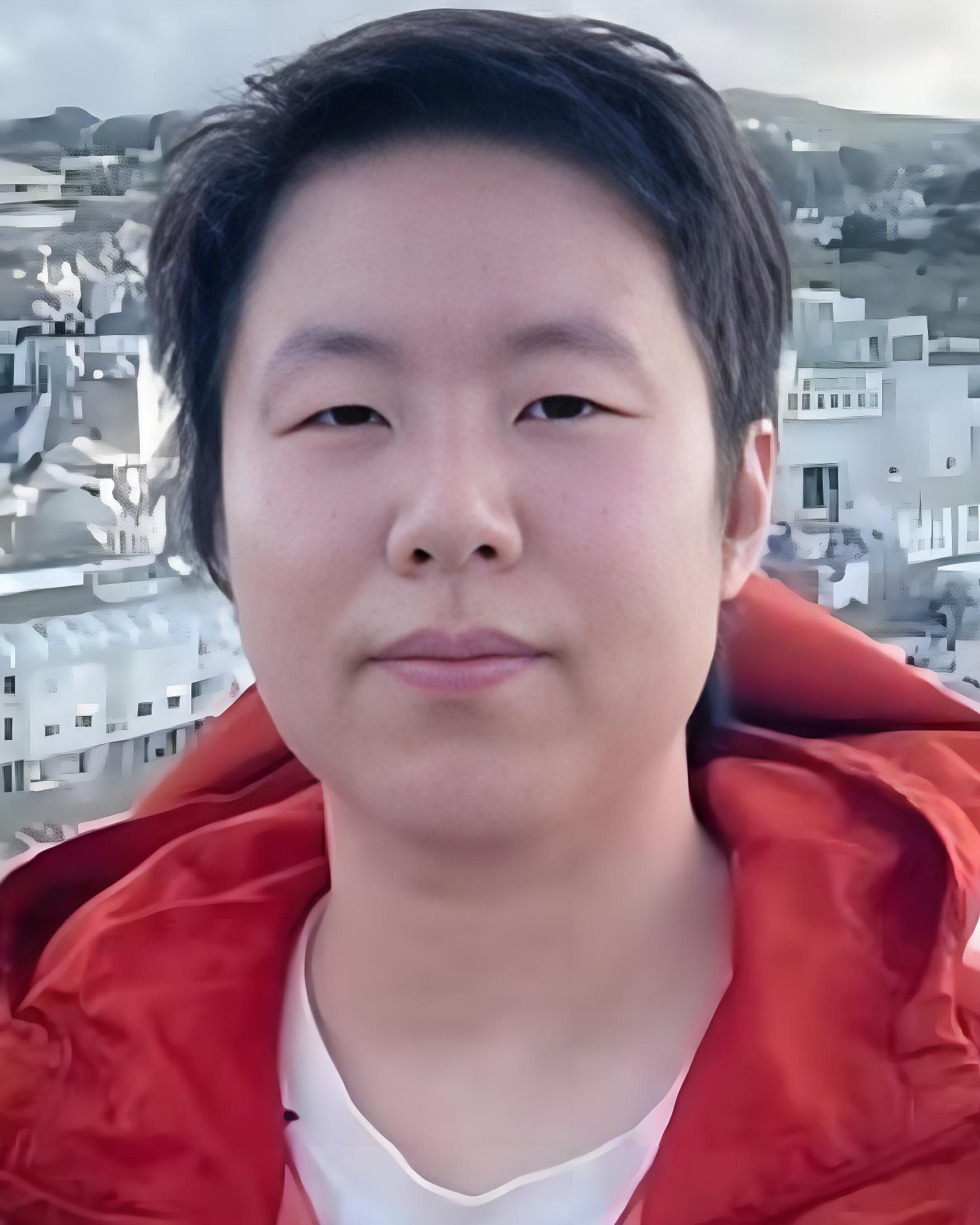}}]{Wei Zhou} (Senior Member, IEEE) received the Ph.D. degree from the University of Science and Technology of China in 2021, jointly with the University of Waterloo, Canada.
They were a Visiting Professor with Dalian University of Technology, a Visiting Scholar with the National Institute of Informatics, Japan, a Research Assistant with Intel, and a Research Intern with Microsoft Research and Alibaba Cloud. They were a Post-Doctoral Fellow with the University of Waterloo. They are currently an Assistant Professor with Cardiff University, U.K. Their research interests include multimedia computing, perceptual image processing, and computational vision.
\end{IEEEbiography}

\begin{IEEEbiography}[{\includegraphics[width=1in,height=1.25in,clip,keepaspectratio]{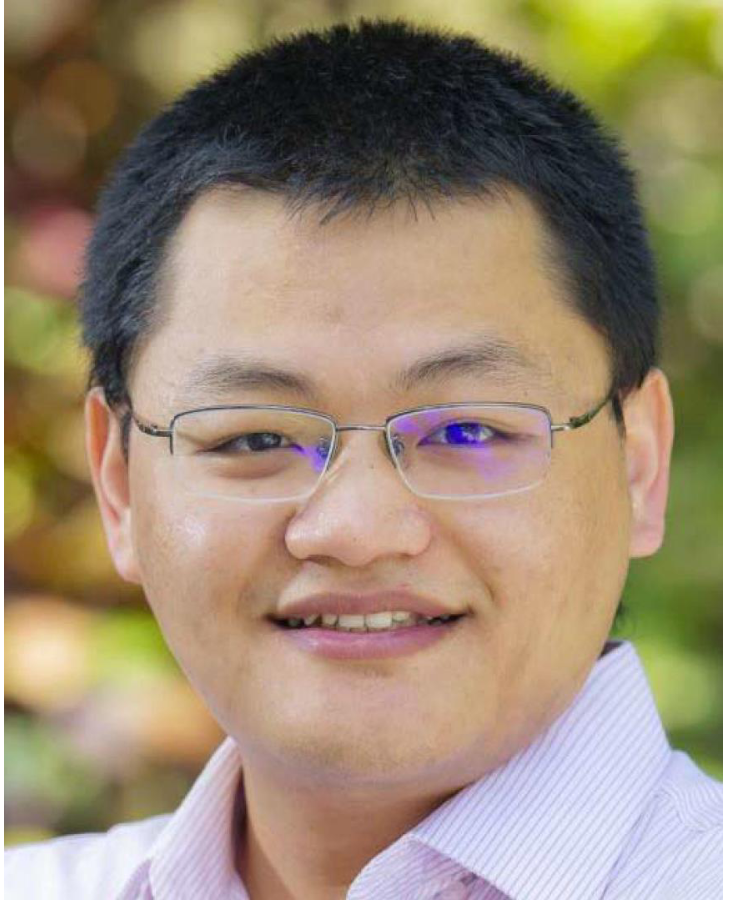}}]{Shiqi Wang} (Senior Member, IEEE) received the Ph.D. degree in computer application technology from Peking University in 2014. He is currently an Professor with the Department of Computer Science, City University of Hong Kong. He has proposed more than 70 technical proposals to ISO/MPEG, ITU-T, and AVS standards. He authored or coauthored more than 300 refereed journal articles/conference papers, including more than 100 IEEE Transactions. His research interests include video compression, image/video quality assessment, video coding for machine, and semantic communication. He served or serves as an Associate Editor for IEEE TIP, TCSVT, TMM, TCyber, Access, and APSIPA Transactions on Signal and Information Processing.
\end{IEEEbiography}

\begin{IEEEbiography}[{\includegraphics[width=1in,height=1.25in,clip,keepaspectratio]{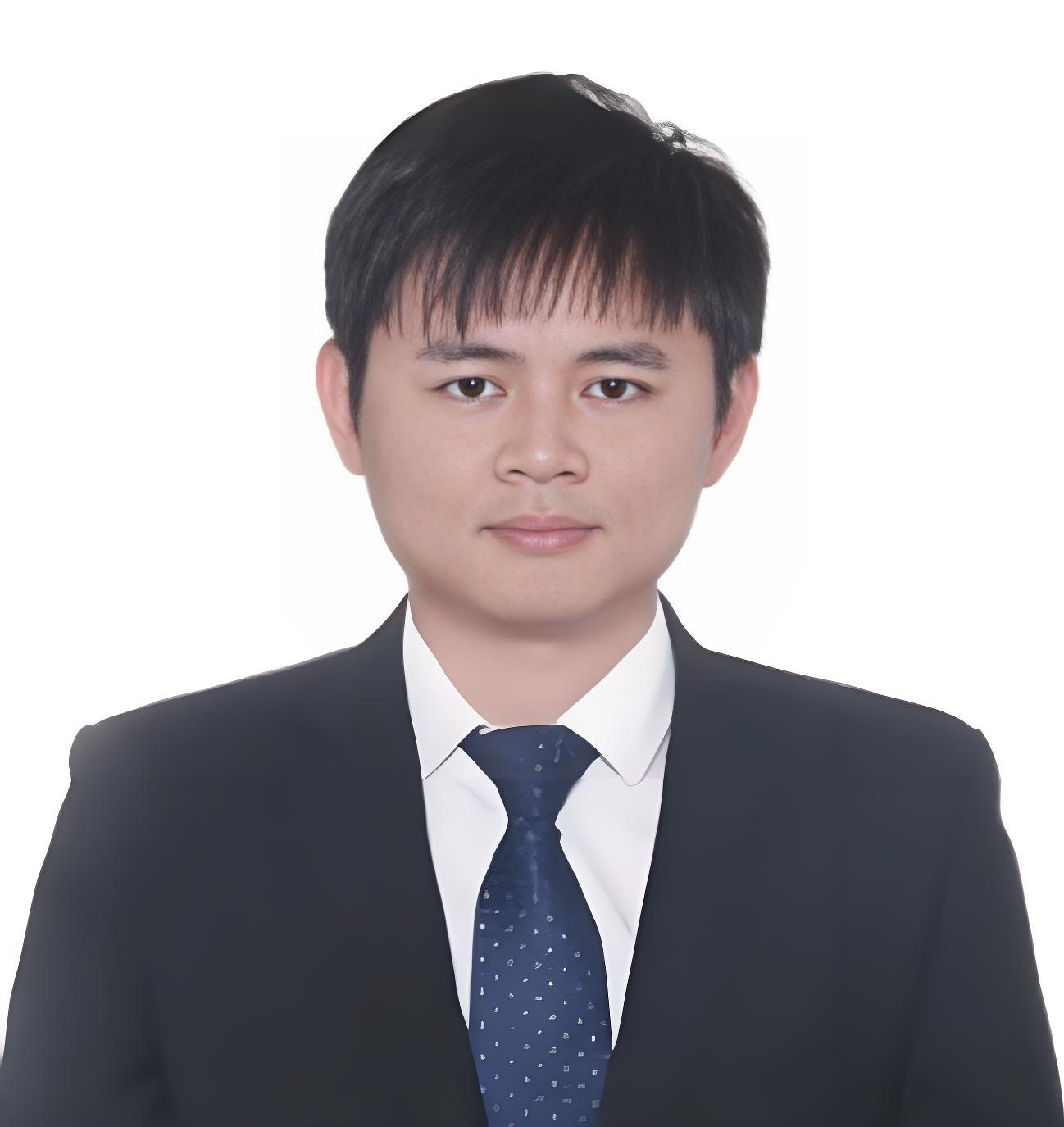}}]{Yuming Fang} (Fellow, IEEE) received the B.E. degree from Sichuan University, Chengdu, China, the M.S. degree from the Beijing University of Technology, Beijing, China, and the Ph.D. degree from Nanyang Technological University, Singapore. He is currently a Professor with the School of Computing and Artificial Intelligence, Jiangxi University of Finance and Economics, Nanchang, China. His research interests include visual attention modeling, visual quality assessment, image retargeting, computer vision, and 3D image/video processing. He serves as an Associate Editor for IEEE TRANSACTIONS ON MULTIMEDIA. He is an Editorial Board of Signal Processing: Image Communication.
\end{IEEEbiography}

\begin{IEEEbiography}[{\includegraphics[width=1in,height=1.25in,clip,keepaspectratio]{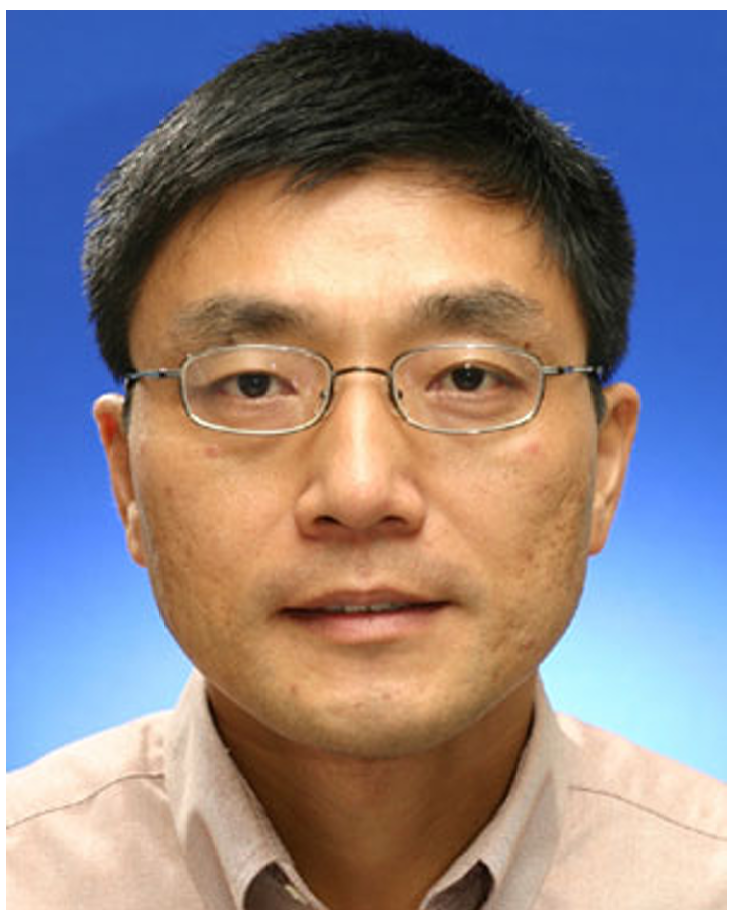}}]{Weisi Lin} (Fellow, IEEE) received the Ph.D. degree from the King’s College, University of London, U.K. He is currently a Professor with the College of Computing and Data Science, Nanyang Technological University. His areas of expertise include image processing, perceptual signal modeling, video compression, and multimedia communication, in which he has published over 200 journal articles, over 230 conference papers, filed seven patents, and authored two books. He has been an invited/panelist/keynote/tutorial speaker at over 20 international conferences. He is a fellow of IET and an Honorary Fellow of Singapore Institute of Engineering Technologists. He has been the Technical Program Chair of IEEE ICME 2013, PCM 2012, QoMEX 2014, and IEEE VCIP 2017. He has been an Associate Editor of IEEE TRANSACTIONS ON IMAGE PROCESSING, IEEE TRANSACTIONS ON CIRCUITS AND SYSTEMS FOR VIDEO TECHNOLOGY, IEEE TRANSACTIONS ON MULTIMEDIA, and IEEE SIGNAL PROCESSING LETTERS. He was a Distinguished Lecturer of Asia-Pacific Signal and Information Processing Association (APSIPA) from 2012 to 2013 and the IEEE Circuits and Systems Society from 2016 to 2017.
\end{IEEEbiography}

\end{document}